\documentclass[sigconf]{acmart}
\pdfoutput=1
\usepackage{float}
\usepackage{stfloats}
\usepackage{amsthm}
\usepackage{hyperref}
\usepackage{url}
\usepackage{graphicx}
\usepackage{wrapfig}
\usepackage{bbm}
\usepackage{subfigure}
\usepackage{caption}
\usepackage{subcaption}
\usepackage{ragged2e} 
\usepackage{enumitem}
\usepackage{balance}
\usepackage{booktabs,makecell, multirow, tabularx}
\AtBeginDocument{%
  }
\newtheorem{proposition}{Proposition}
\newtheorem{lemma}{Lemma}

\copyrightyear{2025}
\acmYear{2025}
\setcopyright{acmlicensed}
\acmConference[WWW '25] {Proceedings of the ACM Web Conference 2025}{April 28--May 2, 2025}{Sydney, NSW, Australia.}
\acmBooktitle{Proceedings of the ACM Web Conference 2025 (WWW '25), April 28--May 2, 2025, Sydney, NSW, Australia}
\acmISBN{979-8-4007-1274-6/25/04}
\acmDOI{10.1145/3696410.3714927}
\newcommand{\ie}{\emph{i.e., }}
\newcommand{\eg}{\emph{e.g., }}

\begin{document}

\title{Uncertainty-Aware Graph Structure Learning}

\author{Shen Han}
\authornotemark[2]
\orcid{0000-0001-6714-5237}
\affiliation{
\department{The State Key Laboratory of Blockchain and Data Security,}
\institution{Zhejiang University}
\city{Hangzhou}
\country{China}}
\email{drhanshen@zju.edu.cn}

\author{Zhiyao Zhou}
\authornotemark[2]
\orcid{0009-0005-9291-169X}
\affiliation{
\department{The State Key Laboratory of Blockchain and Data Security,}
\institution{Zhejiang University}
\city{Hangzhou}
\country{China}}
\email{zjucszzy@zju.edu.cn}
\author{Jiawei Chen}
\authornote{Corresponding author.}
\authornote{College of Computer Science and Technology, Zhejiang University.}
\authornote{Hangzhou High-Tech Zone (Binjiang) Institute of Blockchain and Data Security.}
\orcid{0000-0002-4752-2629}
\affiliation{
\department{The State Key Laboratory of Blockchain and Data Security,}
\institution{Zhejiang University}
\city{Hangzhou}
\country{China}}
\email{sleepyhunt@zju.edu.cn}
\author{Zhezheng Hao}
\orcid{0000-0001-9900-894X}
\affiliation{
\institution{Northwestern Polytechnical University}
\city{Xi'an}
\country{China}}
\email{haozhezheng@outlook.com}

\author{Sheng Zhou}
\orcid{0000-0003-3645-1041}
\affiliation{
\institution{Zhejiang University}
\city{Hangzhou}
\country{China}}
\email{zhousheng_zju@zju.edu.cn}

\author{Gang Wang}
\orcid{0000-0001-6248-1426}
\affiliation{
\institution{Bangsun Technology}
\city{Hangzhou}
\country{China}}
\email{wanggang@bsfit.com.cn}

\author{Yan Feng}
\authornotemark[2]
\orcid{0000-0002-3605-5404}
\affiliation{
\department{The State Key Laboratory of Blockchain and Data Security,}
\institution{Zhejiang University}
\city{Hangzhou}
\country{China}}
\email{fengyan@zju.edu.cn}

\author{Chun Chen}
\authornotemark[2]
\orcid{0000-0002-6198-7481}
\affiliation{
\department{The State Key Laboratory of Blockchain and Data Security,}
\institution{Zhejiang University}
\city{Hangzhou}
\country{China}}
\email{chenc@zju.edu.cn}

\author{Can Wang}
\orcid{0000-0002-5890-4307}
\authornotemark[3]
\affiliation{
\department{The State Key Laboratory of Blockchain and Data Security,}
\institution{Zhejiang University}
\city{Hangzhou}
\country{China}}
\email{wcan@zju.edu.cn}
\renewcommand{\shortauthors}{Shen Han, et al.}
\begin{abstract}
\textit{Graph Neural Networks} (GNNs) have become a prominent approach for learning from graph-structured data. However, their effectiveness can be significantly compromised when the graph structure is suboptimal.
To address this issue, \textit{Graph Structure Learning} (GSL) has emerged as a promising technique that refines node connections adaptively. Nevertheless, we identify two key limitations in existing GSL methods: 1) Most methods primarily focus on node similarity to construct relationships, while overlooking the quality of node information. Blindly connecting low-quality nodes and aggregating their ambiguous information can degrade the performance of other nodes. 2) The constructed graph structures are often constrained to be symmetric, which may limit the model's flexibility and effectiveness.

To overcome these limitations, we propose an \textbf{Uncertainty-aware Graph Structure Learning} (UnGSL) strategy. UnGSL estimates the uncertainty of node information and utilizes it to adjust the strength of directional connections, where the influence of nodes with high uncertainty is adaptively reduced. 
Importantly, UnGSL serves as a plug-in module that can be seamlessly integrated into existing GSL methods with minimal additional computational cost. In our experiments, we implement UnGSL into six representative GSL methods, demonstrating consistent performance improvements.
The code is available at \href{https://github.com/UnHans/UnGSL}{https://github.com/UnHans/UnGSL}.

\end{abstract}



\maketitle

\section{Introduction}
Graph neural networks (GNNs) \cite{kipf2017semisupervised,wu2019simplifying,gilmer2017neural} have demonstrated remarkable performance in tackling graph-structured data.
To date, GNNs have evolved with increasingly sophisticated model architectures \cite{dong2021equivalence,deng2024polynormer,liu2024scalable,wang2024distributionally,chen2024sigformer,chen2024macro} to enhance their capabilities.   
However, these model-centric methods often neglect potential flaws in the underlying graph structure, which can lead to suboptimal performance.  
In practice, graph data frequently exhibit suboptimal characteristics, such as noisy connections and incomplete information, due to the inherent complexities and inconsistencies in data collection \cite{zhou2023opengsl,um2023confidencebased}.

\begin{figure}[t]
    \setlength{\abovecaptionskip}{5pt}
    \subfigure[Cora]{
       \centering
       \includegraphics[width=0.45\linewidth,trim = 55 30 -10 10]{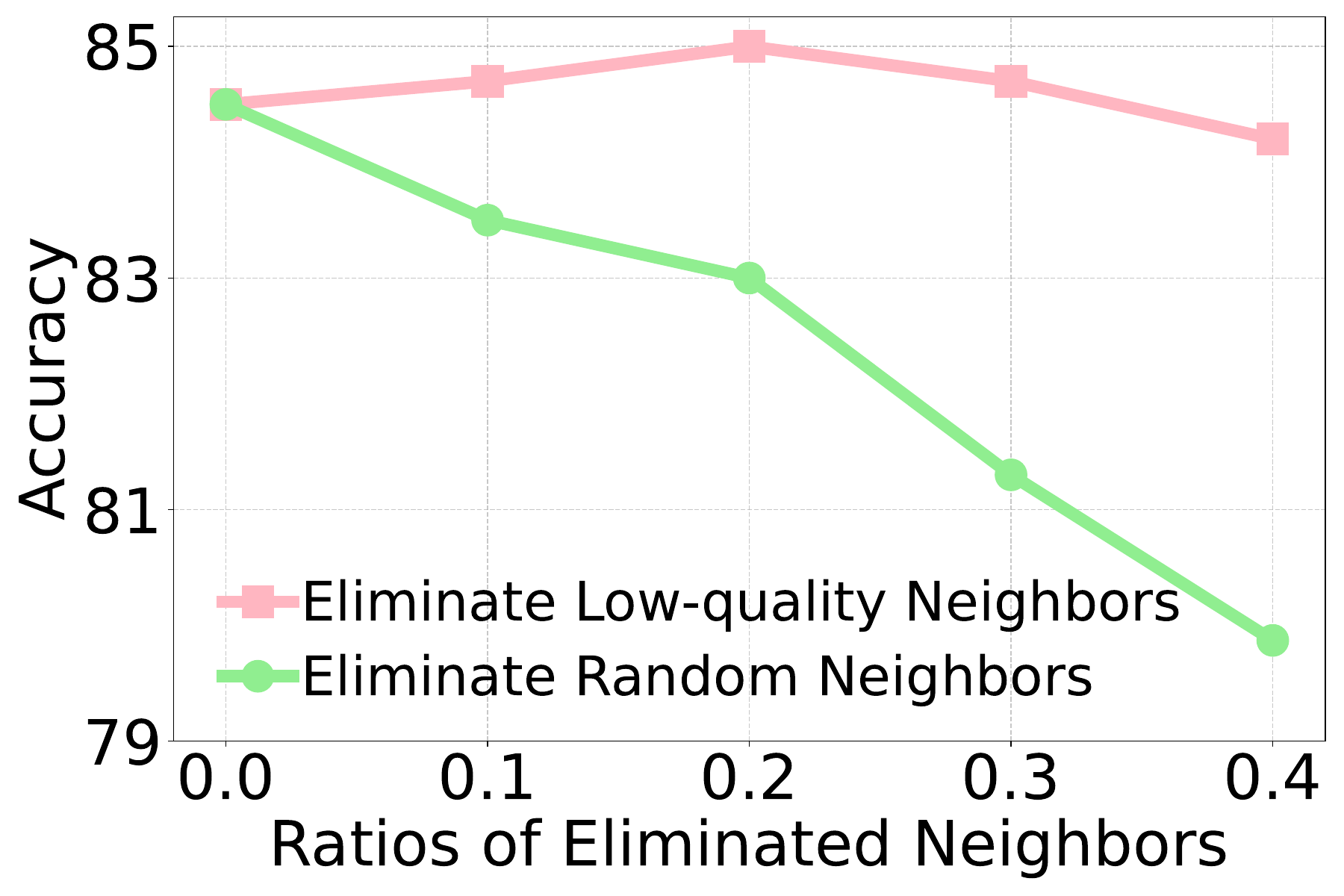}
    }
    \subfigure[Citeseer]{
        \centering
        \includegraphics[width=0.45\linewidth,trim = 35 20 20 10]{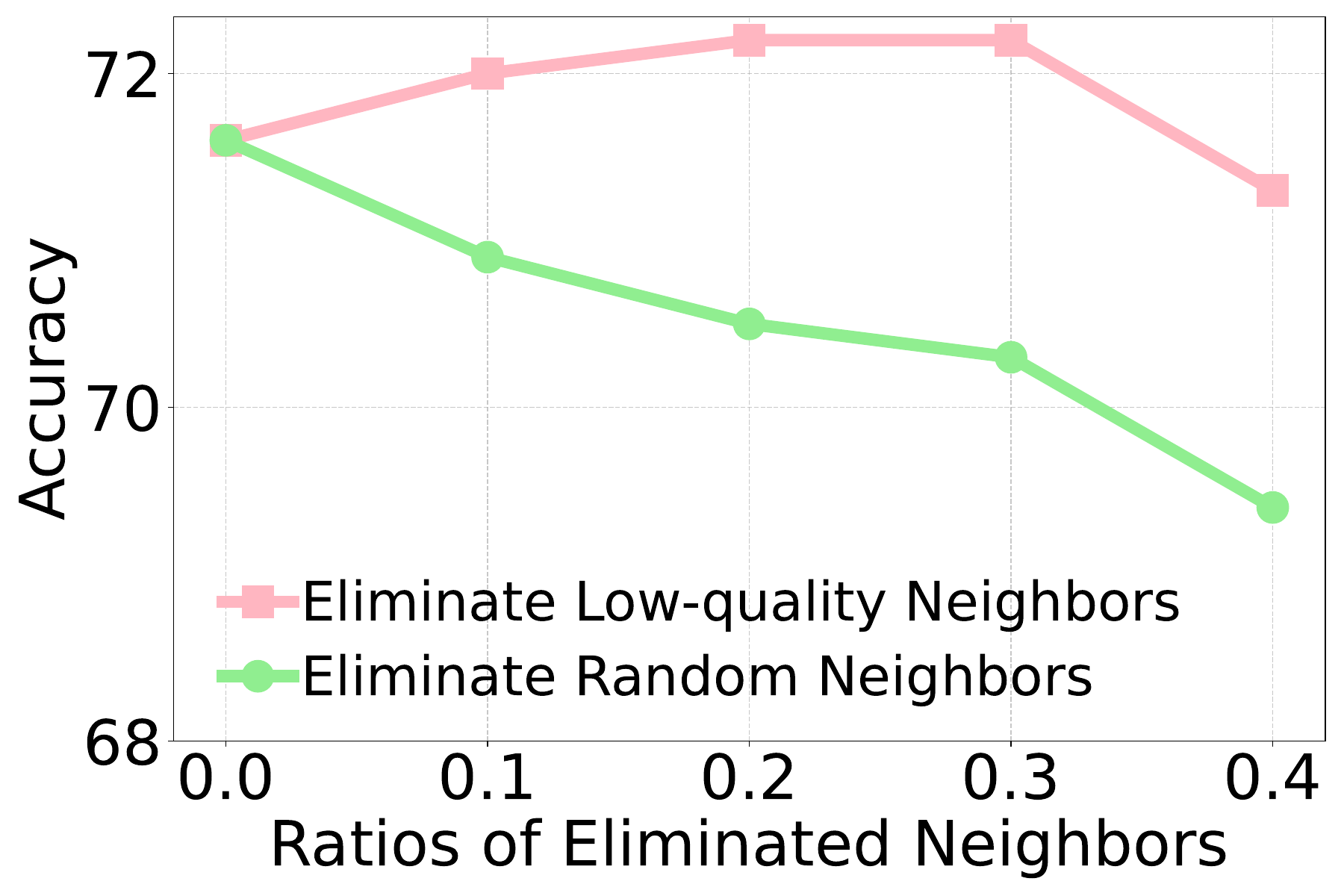} 
    }

    \caption{Performance varies with the ratios of eliminated neighbors on Cora and Citeseer datasets. Here we first constructed a graph based on GRCN \cite{yu2021graph}. We then eliminate a certain ratio of neighbors with the highest entropy for each node, and evaluate the performance of GCN on such a pruned graph. 
    {For comparison, we also report the performance under random elimination.}
    }

    \setlength{\abovecaptionskip}{-3.cm}
    \label{fig1}
    \vspace{-15pt}
    \Description{}
\end{figure}

To address these issues, Graph Structure Learning (GSL) \cite{jin2020graph,chen2020iterative,fatemi2021slaps,zhang2021hierarchical,lu2024latent}, a data-centric approach, has garnered increasing attention. 
Beyond learning node representations with GNNs, GSL learns to refine node connections and edge weights. 
This approach has been shown to effectively enhance the accuracy of GNNs on downstream tasks while improving their resilience to topological perturbations \cite{zhou2023opengsl,li2023gslb}.   
Early work on GSL directly treated the graph structure (\ie the adjacency matrix) as learnable parameters. 
However, due to the large parameter space, these strategies often incur substantial computational overhead and are difficult to train effectively \cite{franceschi2019learning,jin2020graph}. 
More recently, research has shifted towards embedding-based GSL \cite{chen2020iterative,liu2022compact,wang2023prose,in2024self},  which constructs the adjacency matrix based on the similarity of node embeddings. 
Various similarity metrics, such as cosine similarity \cite{wang2023prose,in2024self} or neural networks \cite{liu2022compact}, have been employed. 
These methods aim to increase graph homophily and typically achieve state-of-the-art performance, as nodes with similar features (or embeddings) are more likely to be connected.

\begin{figure}[t]
\centering
\includegraphics[width=0.8\linewidth, trim = 70 20 70 -10]{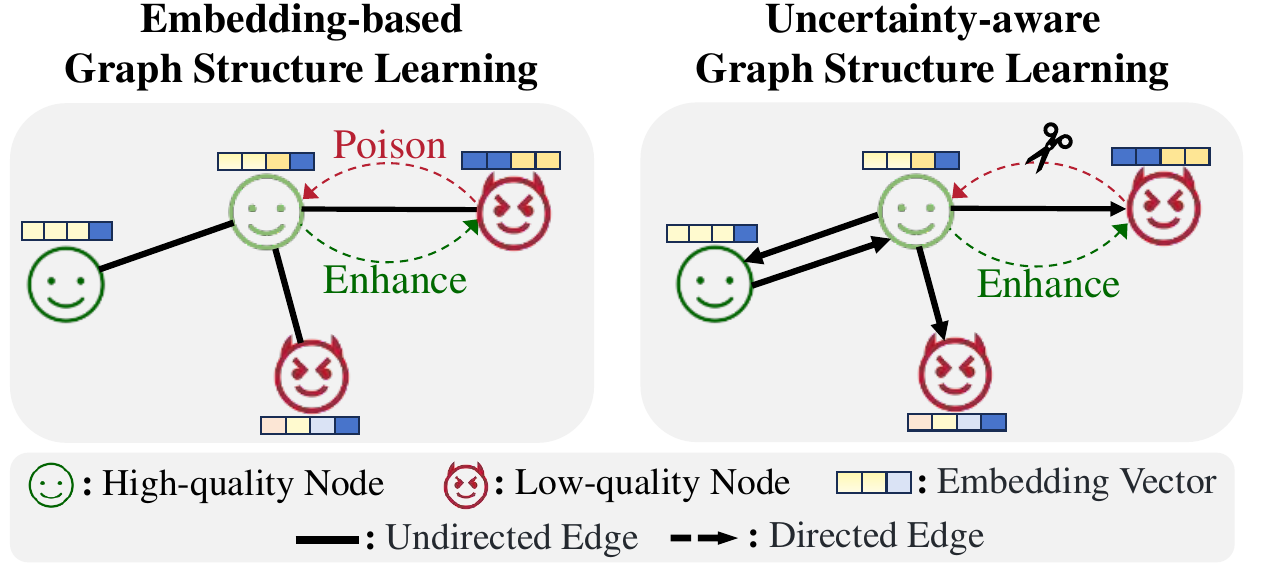}
\caption{Illustration of how our UnGSL differs from existing embedding-based GSL methods. Existing GSL learns symmetric relationships, leading to a dilemma when managing connections between high-quality and low-quality nodes.
In contrast, UnGSL learns asymmetric relationships, allowing low-quality nodes to benefit from high-quality nodes while mitigating the negative influence of low-quality nodes.}

\label{fig2}
\vspace{-15pt}
\Description{}
\end{figure}

Despite their success, we identify two key limitations in these embedding-based GSL methods:
\begin{itemize} [leftmargin=1.0em]
    \item \textbf{These methods mainly rely on embedding similarity for graph construction while neglecting the quality of node information.} 
    Given the critical role of edges in GNNs as conduits for information propagation, it is essential to evaluate the quality of the information being propagated.  
    Aggregating unclear or ambiguous information from neighbor nodes can disrupt the embedding learning of the target node. 
    Constructing connections based solely on node similarity, without assessing the quality of the node’s information, may lead to suboptimal performance. 
    To empirically validate this point, we conduct a simple experiment using a representative GSL method (GRCN~\cite{yu2021graph}). 
    As shown in Figure \ref{fig1}, removing a certain proportion of neighbors with the highest entropy results in a significant performance gain. 
    \item \textbf{These methods often generate symmetric graph structures, which can hinder their effectiveness.} Current embedding-based methods tend to construct symmetric relationships between nodes, implying that both nodes exert an equal and bidirectional influence during the GNN learning process.  
    This imposed symmetry can constrain the model's flexibility and capacity, particularly when the connected nodes differ in quality. For example, consider a scenario where a high-quality node is linked to a low-quality node (refer to Figure \ref{fig2}).  While the high-quality node can provide valuable information that greatly benefits the low-quality node, the reverse influence from the low-quality node may have a negative impact on the high-quality node. 
    Symmetric relationships fail to account for this disparity, leading to a dilemma in the learning process.
    This inspires us to explore asymmetric structure learning.
    By modeling directional relationships separately, we allow the low-quality node to benefit from the high-quality node’s information while reducing the adverse influence in the opposite direction, thus protecting the high-quality node from negative effects. Although a few studies \cite{song2022towards,song2024optimal,lu2024latent} have begun to explore asymmetric graph structure learning, they typically restrict the asymmetry to relations between labeled and unlabeled nodes, overlooking the richer relationships between the vast majority of unlabeled nodes.
\end{itemize}

To overcome these limitations, we propose an uncertainty-aware graph structure learning (UnGSL) method that considers nodes' information quality to learn an asymmetric graph structure. 
UnGSL directly utilizes the uncertainty (Shannon Entropy \cite{abdar2021review}) of the node in classification to indicate the node's information quality, and conducts theoretical analyses to demonstrate that blindly aggregating information from high-uncertainty nodes would lift the lower bound of uncertainty for the target node.
Building on this, UnGSL leverages a learnable node-wise threshold to differentiate low-quality neighbors from high-quality ones, and adaptively reduces directional edge weights from those low-quality neighbors. 
Notably, our UnGSL is simple and can be easily incorporated into various embedding-based GSL methods, boosting their performance with minor extra computational overhead.

In summary, this work makes the following contributions:
\begin{itemize} [leftmargin=1.0em]
\item We highlight the necessity of modeling a node's uncertainty in graph structure learning, and theoretically demonstrate that the uncertainty of a node after GNN layer is positively correlated with those of its neighbors.
\item We propose a simple yet novel uncertainty-aware graph structure learning strategy (UnGSL), which can be seamlessly integrated with various embedding-based GSL models to mitigate the directional impact of high-uncertainty nodes.
\item We conduct extensive experiments to demonstrate that UnGSL can consistently boost existing embedding-based GSL models across seven benchmark datasets, with an average performance increase of 2.07\%.
\end{itemize}

\section{Preliminaries}
\label{sec2}
In this section, we introduce  basic notations and  background on GNNs and GSL methods.

Consider the graph $\mathcal{G}=(\mathcal{V}, \mathcal{E}, \mathbf{A}, \mathbf{X})$, where $\mathcal{V}$ represents a set of $n$ nodes $\{v_1,...,v_n\}$ and $\mathcal{E}$ represents the set of edges. Let $\mathbf{A}$ be the initial adjacency matrix of the graph, where $\mathbf{A}_{ij}=1$ if an edge exists between node $v_i$ and $v_j$; otherwise, $\mathbf{A}_{ij}=0$. The matrix $\mathbf{X}=[x_1,...,x_n]\in \mathbb{R}^{n \times d}$ represents the node feature matrix, where each column $x_i$ corresponds to the feature vector of node $v_i$. Let 
$\mathbf{D}$ denote the diagonal degree matrix defined as $\mathbf{D}_{ii}=1+\sum_{j}\mathbf{A}_{ij}$; and 
$\hat{\mathbf{A}}$ denotes the normalized adjacency matrix with self-loop, \ie $\hat{\mathbf{A}}=\mathbf{D}^{-\frac{1}{2}} (\mathbf{A}+\mathbf{I}) \mathbf{D}^{-\frac{1}{2}}$ for symmetric normalization or $\hat{\mathbf{A}}=\mathbf{D}^{-1} (\mathbf{A}+\mathbf{I})$ for row normalization. 
\subsection{Graph Neural Networks}
Graph Neural Networks (GNNs)  have become a prominent approach for learning from graph-structured data, where node representations are learned by iteratively aggregating and transforming information from neighboring nodes. 
In recent years, various designs for the aggregation and transformation processes have given rise to different GNN models\cite{kipf2017semisupervised,veličković2018graph,wu2019simplifying}. 
Among these, the Graph Convolutional Network (GCN) \cite{kipf2017semisupervised} stands out as one of the most widely adopted and influential architectures. 
The operation at the $l$-th layer in GCN can be formulated as:
\begin{equation}
    \mathbf{Z}^{(l)}=\sigma(\hat{\mathbf{A}}\mathbf{Z}^{(l-1)}\mathbf{W}^{(l)}),
\end{equation}
where $\sigma( \cdot )$ denotes the activation function, $\mathbf{W}^{(l)}$ is a learnable parameter matrix used to transform the node features. 

Given the critical role of the graph structure in GNNs, which determines the sources of information aggregation, ensuring the quality of graph structure is of paramount importance. Recent work demonstrates that suboptimal graph structures,  even with the introduction of a small percentage of noisy edges or topological perturbations (\eg 10\%), can significantly degrade the performance of GNNs (\eg 25\%) \cite{zhou2023opengsl,li2023gslb}.

\subsection{Graph Structure Learning}
Graph Structure Learning (GSL) aims to enhance the accuracy and robustness of GNNs by learning a refined adjacency matrix $\mathbf{S}$, where $\mathbf{S}_{ij}$ denotes the edge weight between node $v_{i}$ and node $v_{j}$. 
For convenience, we also use $\mathbf{S}_{ij}$ to indicate whether there exists a edge between node $v_i$ and $v_j$.
When $\mathbf{S}_{ij}>0$, it indicates the presence of a directed edge from $v_{j}$ to $v_{i}$
, with the edge weight equal to $\mathbf{S}_{ij}$. 
Conversely,  $\mathbf{S}_{ij}=0$ indicates that no edge exists.

The quality of the learned adjacency matrix is often examined on the downstream tasks. 
Conventionally, $\mathbf{S}$ is fed into a GNN encoder to obtain node representations, which are used to evaluate performance on downstream tasks.
The objective function of most GSL methods can be generally formulated as:
\begin{equation}
    \mathcal{L} =\mathcal{L}_{{Task}}(\mathbf{Z},\mathbf{Y})+\lambda\mathcal{L}_{{Reg}}(\mathbf{Z},  
    \mathbf{S}),
    \label{eq_summary_gsl}
\end{equation}
where $\mathbf{Z}=\text{GNN}(\mathbf{S},\mathbf{X})$ is the node embedding matrix, $\mathcal{L}_{\text{Task}}$ aims to utilize the labels $\mathbf{Y}$ as a supervised signal to optimize both the GNN encoder and the adjacency matrix $\mathbf{S}$.
The regularization term $\mathcal{L}_{\text{Reg}}$ are introduced by existing GSL methods to achieve diverse desirable properties in the learned adjacency matrix $\mathbf{S}$, such as sparsity \cite{jin2020graph}, feature smoothness \cite{song2024optimal}, and connectivity \cite{yu2021graph} in the graph structure.
For example, PROGNN \cite{jin2020graph} employs $l_{1}$ norm penalization on the learned adjacency matrix to enhance the sparsity of the graph structure.
$\lambda$ is a trade-off hyperparameter.

Traditional GSL methods \cite{franceschi2019learning,jin2020graph} that treat each edge $\mathbf{S}_{ij}$ as a learnable parameter often suffer from significant computational overhead and are challenging to train efficiently.
Recent research has focused on embedding-based GSL methods \cite{chen2020iterative,liu2022compact,wang2023prose,in2024self} , which employ an auxiliary GNN encoder to extract node embeddings from the initial graph and estimate embedding similarities as edge weights:
\begin{equation}
\label{embed}
    \mathbf{E} = \text{GNN}(\mathbf{A},\mathbf{X}),
\end{equation}
\begin{equation}
\label{sim}
    \mathbf{S}_{ij} = \mathbf{S}_{ji} = \phi(\mathbf{E}_{i},\mathbf{E}_j).
\end{equation}
Here $\mathbf{E}$ is the node embedding matrix,  which differs from $\mathbf{Z}$ as it is specifically used for constructing the graph structure.
Note that existing works generally utilize different GNN architectures and input structure compared with those used to compute the embedding $\mathbf{Z}$ when constructing the embedding $\mathbf{E}$.
$\phi(\cdot)$ is a metric function (\ie cosine similarity) used to calculate the similarity between nodes.

Although the structure modeling paradigm in Equation (\ref{sim}) is widely adopted in GSL models, it suffers from two key limitations:
\begin{itemize}[leftmargin=1.0em] 
    \item  \textbf{This paradigm relies on embedding similarity while neglecting the quality of node information.}
    {Only semantic similarities between embeddings $\mathbf{E}_i$  and $\mathbf{E}_j$ are considered in this structure modeling paradigm, while the varying uncertainties of them, which reflect their information quality, are neglected.
    This may undermine the quality of embeddings of target nodes when aggregating inferior information from low-quality neighbors (as validated by the preliminary experiment in the Introduction) .}    
    \item \textbf{This paradigm constraining the graph to be symmetric, which potentially hinder effectiveness of GSL models. }
    The pair-wise similarity  constrains the learned edge between $v_i$ and $v_j$ to be bidirectional, overlooking their unequal influence due to varying information quality.
    For example, if $\mathbf{E}_i$ contains higher-quality information than $\mathbf{E}_j$, the constructed edge $\mathbf{S}_{ji}$ can provide valuable information that greatly benefits the low-quality node $v_j$ while the edge $\mathbf{S}_{ij}$ propagates inferior information to poison the embedding of $v_i$.
    By modeling directional relationships separately, we enable the low-quality node to benefit from the high-quality node's information while mitigating the negative impact in the reverse direction.
\end{itemize}

Although a few works \cite{song2022towards,song2024optimal,lu2024latent} have been proposed to learn asymmetric graphs by generating directed edges $\mathbf{S}_{ij}$ from the labeled node $v_j$ to unlabeled node $v_i$ and constraining $\mathbf{S}_{ji}=0$, which facilitates the propagation of label information and avoids introducing inconsistency to labeled nodes.
However, these methods completely rely  on annotated labels and fail to learn reasonable asymmetric connections between the vast majority of unlabeled nodes.

Given the flaws of existing methods, we argue for the necessity of incorporating node uncertainty into graph structure learning to learn an optimal asymmetric structure.
We propose the uncertainty-aware graph structure learning (UnGSL) method to enhance GSL models, which leverage learnable node-wise thresholds to identify high-uncertainty neighbors and adaptively reduce directional edge weights from those low-quality neighbors.
\section{Methodology}
\label{sec3}
{In this section, we first conduct theoretical and empirical analyses to demonstrate the detrimental impact of neighbors with high uncertainty levels on GNN learning (Subsection~\ref{sec3.1}). 
We then present the proposed uncertainty-aware graph structure learning method in detail (Subsection ~\ref{sec3.2}).}

\begin{figure}[t]

       \centering
       \includegraphics[width=0.8\linewidth,trim = 35 30 -10 -5]{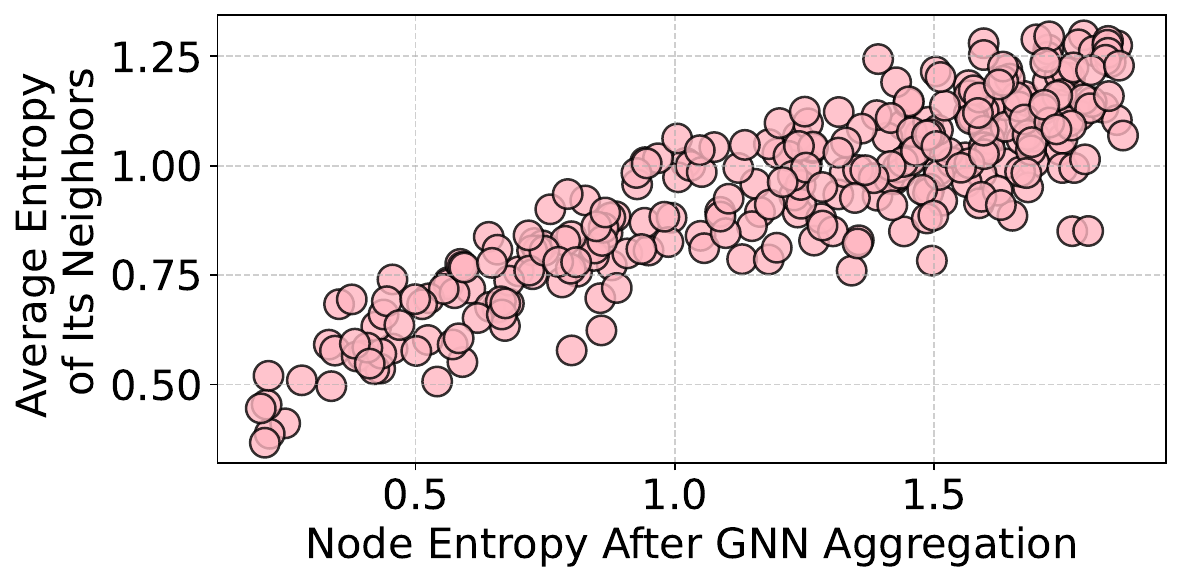}

    \caption{Visualization of node entropy after GNN aggregation  (i.e., $u_{i}$) alongside the average entropy of its neighbors  (i.e., $\sum_{v_j \in \mathcal{N}{(v_i)}}\mathbf{\hat{A}}_{ij} u_{j}^{\prime}$) on Cora dataset.\label{Figure2}}
    \setlength{\belowcaptionskip}{-15.cm}
    \vspace{-15pt}
    \Description{}
\end{figure}

\subsection{Analyses on the Impact of Neighbor Uncertainty}
\label{sec3.1}
Aggregating unclear or ambiguous information can intuitively disrupt the learning process of target nodes, thereby negatively affecting the performance of Graph Neural Networks (GNNs). In this section, we aim to conduct both theoretical and empirical analyses to substantiate this claim. To begin, we introduce several formal concepts to facilitate these analyses.

\textbf{Semi-supervised Node Classification Task.} For convenience, we refer to recent analytical work on GNNs \cite{ma2022is} and focus our theoretical analysis on the semi-supervised node classification task, which is the most common and widely studied scenario. Nevertheless, at the end of this section, we will also discuss how our method can be adapted to the unsupervised learning scenario.  
Following the definitions in \cite{ma2022is}, we consider a $K$-class classification problem and employ a linear classification model. 
Formally, the classification logits can be expressed as:
\begin{equation}
    \mathbf{O}=\mathbf{D^{-1}(A+I)X}\mathbf{W},
\end{equation}
where $\mathbf{W} \in \mathbb{R}^{d \times K}$ is the linear classification matrix.
Assume the logit in matrix $\mathbf{O}$ are bounded by the scalar 1, \ie max $|\mathbf{O}_{ij}| < 1$.
From a local perspective for node $v_i$, its probabilities can be formulated as :
\begin{equation}
    p_{i} = \frac{\mathbf{O}_{i} +  \mathbf{1}_{K} }
    {\sum_{j=1}^{K}(\mathbf{O}_{ij} + 1)},
\label{eq3}
\end{equation}
where $\mathbf{1}_{K}$ is a $K$-dimensional all-ones vector.
Here we simply omit the nonlinear exponential function in the softmax, given that our primary focus is on the impact of aggregation on node uncertainty and the exponential function mainly serves to generate positive values.  
This simplification has been adopted in prior work \cite{zhao2020uncertainty,yu2023uncertainty}, and it can also be considered a first-order Taylor approximation.


\textbf{Uncertainty Estimation via Entropy.}
Entropy measures the information uncertainty within a probability distribution \cite{abdar2021review}.
For node classification tasks, the entropy of the classification probabilities reflects the classifier's certainty in assigning the node representation to a specific class.
A high-entropy node indicates that its representation carries significant uncertainty, making it challenging for the classifier to reach a confident decision. Aggregating information from such nodes can poison the target node's representation, hindering the generation of accurate predictions.
Given probabilities $p_i$ of node $v_i$, its entropy is defined as:
\begin{equation}
\label{eq7}
    u_{i} = - \sum_{k=1}^{K} p_{ik} \text{log}(p_{ik}).  
\end{equation}
We further discuss  the uncertainty metric for unsupervised learning scenario at the end of this section.

Formally, to demonstrate the impact of the uncertainty of neighbors along GNNs, we have the following proposition, with detailed proof provided in Appendix \ref{proof}: 
\begin{proposition}
\label{prop1}
Define the logits of the initial node feature matrix as $\mathbf{O}^{\prime} = \mathbf{X}\mathbf{W}$.
For a given node $v_i$, let $u_i$ denote its entropy after GNN aggregation. 
$\forall v_j \in \mathcal{N}{(v_i)}$, let $u_j^{\prime}$ denote the entropy of its initial features.
Then the entropy of $v_i$ and the entropy of  $v_j \in \mathcal{N}{(v_i)} $ satisfy the following inequality:
\begin{equation}
     u_i \geq \sum_{v_j \in \mathcal{N}{(v_i)}}\eta_{j} u_{j}^{\prime},
\end{equation}
where 
\begin{equation}
\eta_{j}=\frac{
\sum_{k=1}^{K}\hat{\mathbf{A}}_{ij}{(O_{jk}^{\prime}+1})}
{ \sum_{v_j \in \mathcal{N}{(v_i)}}
\sum_{k=1}^{K}\hat{\mathbf{A}}_{ij}(O_{jk}^{\prime} + 1) }.
\end{equation}
For coefficient $\eta_{j}$, we have $ 0<\eta_{j} <1$ and $\sum_{v_j \in \mathcal{N}{(v_i)}}\eta_{j} =1$.
\label{proposition 1}
\end{proposition}
 
\textbf{Discussion.}
Proposition \ref{prop1} establishes the bounded relationship between the uncertainty of a targe node after aggregation and the uncertainty of its neighbors. 
It suggests blindly aggregating information from high-uncertainty nodes would lift the lower bound of the uncertainty for the target node. 
Removing or mitigating the impact of these high-uncertainty nodes, may significantly decrease the uncertainty of the target node, leading to performance gains.

\textbf{Empirical Analyses.}
We conduct simple experiments
to empirically demonstrate that aggregating neighbors with higher uncertainty would result in a high-uncertainty node.
Specifically, we train the model with 1-layer GCN and linear classifier on given datasets. Then we visualize the entropy of a node after aggregation (\ie $u_{i}$) alongside the average entropy of its neighbors (\ie $\sum_{v_j \in \mathcal{N}{(v_i)}}\mathbf{\hat{A}}_{ij} u_{j}^{\prime}$).
As shown in Figure \ref{Figure2}, we observe a strong positive correlation between the entropy of a node after aggregation and the average entropy of its neighbors, which substantiates our proposition (results on more datasets are provided in Appendix \ref{ADDC1}).

The above analyses clearly illustrates the impact of node uncertainty in aggregation process in GNNs. 
Blindly connecting and aggregating information from nodes with high uncertainty may undermine the performance of the nodes themselves. 
It is therefore important to consider node uncertainty in graph structure learning to learn a reasonable asymmetric structure.
Specifically,
one can prevent a node falling into a high-uncertainty region by weakening its connections to neighbors with high uncertainty. 
Meanwhile, it can receive more stable information via strengthened connections with low-uncertainty nodes.

\subsection{Uncertainty-aware Graph Structure Learning}
\label{sec3.2}
Given the importance of considering uncertainty in graph structure learning, we propose the simple yet novel uncertainty-aware graph structure learning (UnGSL) method that leverages learnable node-wise thresholds to distinguish low-quality neighbors from high-quality ones and adaptively refines edges based on their uncertainty levels.
Beyond similarity, we introduce uncertainty to guide the learning of the refined adjacency matrix.
Specifically, the uncertainty estimated through entropy is transformed into confidence scores.
Afterwards, UnGSL applies learnable node-wise thresholds to split neighbors with different levels of confidence into two groups (\ie high-confidence and low confidence) for each node. 
Finally, it amplifies edge weights from confident neighbors while reducing edge weights from uncertain ones.
In summary, we refine the generation of adjacent matrix of existing GSL from Equation (\ref{sim}) into Equation (\ref{eq10}):
{
\begin{equation}
\label{eq10}
    \hat{\mathbf{S}}_{ij} = \mathbf{S}_{ij} \cdot \psi({e^{-u_{j}} -{{\varepsilon}_{i}}} ),
\end{equation}
}
Here $u_{j}$ is entropy of node $v_{j}$ estimated during pretraining stage, $\varepsilon_{i}$ is the learnable node-wise threshold of node $v_{i}$, 
$\psi(\cdot)$ is an activate function,
$\mathbf{S}$ is the generated adjacency matrix of the GSL model.

Equation (\ref{eq10}) first normalizes node uncertainty into confidence within the interval $(0,1)$ , and then leverages node-wise learnable thresholds $\boldsymbol{\varepsilon}$ to distinguish low-confidence neighbors (\ie ${e}^{-u_{j}} < {\varepsilon_{i}}$) from high-confidence ones (\ie ${e}^{-u_{j}} \geq {\varepsilon_{i}}$ ) and adaptively refines the corresponding edges using the following activation function:
\begin{equation}
\label{eq11}
    \psi(x) = \left\{
    \begin{array}{lr}
         \tau \cdot s(x),  &{x \geq 0,}\\
        {\beta}, &{x < 0,}
    \end{array} 
    \right.
\end{equation}
where $s(\cdot)$ is the sigmoid function, $\tau$ is a hyperparameter amplifies the edge weights from high-confidence neighbors, $\beta$ is a hyperparameter that controls the reduction of edge weights from low-confidence neighbors.

For the training of UnGSL, we employ the same loss function (\ie Equation (\ref{eq_summary_gsl})) and the same training procedure as the original GSL model, except we modify the generation of the adjacency matrix from Equation (\ref{sim}) to Equation (\ref{eq10}), \ie using 
$\hat{\mathbf{S}}$ to replace the generated adjacency matrix $\mathbf{S}$ in existing GSL methods. 
Therefore, UnGSL can be seamlessly integrated into existing GSL methods with minimal additional computational cost.

\textbf{Implementation Details.}
We begin by selecting a specific GSL method that UnGSL aims to enhance , pretraining it to obtain the classifier and estimating uncertainty using entropy (Equation (\ref{eq7})).
Next, the chosen GSL method is re-trained, during which Equation (\ref{eq10}) is employed to generate the adjacency matrix. 
Finally, the learned graph structure is leveraged for downstream applications.
\subsection{Discussion}
Based on the above formulation, we next discuss several key advantages of UnGSL:

\textbf{Uncertainty-aware.}
UnGSL considers the information quality of nodes during the structure modeling process, facilitating the learning of an optimal graph that mitigates the negative impact of high-uncertainty neighbors in GNN learning.

\textbf{Model-agnostic.}
UnGSL refines the learned adjacency matrix $\mathbf{S}$ by utilizing uncertainty-aware weights. 
In essence, UnGSL just replaces the calculation of the adjacency matrix from Equation (\ref{sim}) to Equation (\ref{eq10}) for GSL methods. 
Therefore, UnGSL can be seamlessly integrated into existing GSL methods to further enhance their ability to learn graphs, improving the performance of GNNs on downstream tasks. 
Comprehensive experiments supporting this can be found in Subsection~\ref{RQ1} and Subsection \ref{RQ3}.

\textbf{Asymmetric Graph Structure.}
UnGSL proposes learning an
asymmetric graph, where the edge weights between nodes differ based on their uncertainty levels.
Specifically, UnGSL weakens edges from uncertain nodes to confident nodes, mitigating the impact of inferior information. Meanwhile it enhances the edge in the opposite direction to improve the representations of uncertain nodes.

Notably, several methods that construct directed edges from labeled nodes to unlabeled nodes~\cite{song2024optimal,lu2024latent} can be viewed as specific cases of UnGSL, as the embeddings of labeled nodes are directly optimized during training and therefore more likely to exhibit lower uncertainty.
In contrast, UnGSL models the asymmetric relationships among the predominantly unlabeled nodes, leading to superior performance. We further empirically validate this in Subsection~\ref{RQ1}.

\textbf{Efficiency.}
UnGSL contains only $n$ learnable parameters and refines the existing edges of the given graph without generating new ones. 
The additional operations introduced by UnGSL have a complexity of $O(n+m)$, where $n$ and $m$ denote the number of nodes and edges in the graph.
Therefore, UnGSL imposes minimal computational costs on the base GSL models.
We provide empirical evidence to demonstrate the efficiency of UnGSL in Subsection~\ref{seceffiency}.

\textbf{Adaptive to Unsupervised Scenarios.}
Despite the mainstream focus of GSL research on supervised learning, our approach can be generalized to unsupervised GSL scenarios. The main challenge lies in determining an appropriate uncertainty metric for unsupervised GSL method.
We suggest employing the self-supervised structure learning loss 
(\ie node contrastive learning loss \cite{liu2022towards}) as a proxy for uncertainty, which can be formulated as:
\begin{equation}
\label{eq13}
    u_{i} = \frac{1}{2}(l(z_{i},{\tilde{z}}_{i})+ l({\tilde{z}}_{i},z_{i}),
\end{equation}
where 
\begin{equation}
l(z_{i},{\tilde{z}}_{i}) = - \log\frac{e^{\text{sim}(\mathrm{z}_{i},\mathrm{\tilde{z}}_{i})/t}}{\sum_{k=1}^n e^{\text{sim}(\mathrm{z}_{i},\mathrm{\tilde{z}}_{k})/t}}.
\end{equation}
Here $\text{sim}(\cdot)$ is the cosine similarity function.
$\mathrm{z}_{i}$ and $\mathrm{\tilde{z}}_{i}$ are the embeddings of node $v_{i}$ learned by GNN from target graph and augmented graph, respectively. 
$\sum_{k=1}^n e^{\text{sim}(\mathrm{z}_{i},\mathrm{\tilde{z}}_{k})/t}$ is cumulative similarity between $z_{i}$ and embeddings of different nodes in the augmented graph. 
$t$ is the temperature parameter.
The GCL loss can measure the invariance of node representations to feature or structure perturbations, where this invariance can be interpreted as the uncertainty of nodes with respect to their original features and structure.


\section{Experiments}
\label{sec4}
In this section, we conduct experiments to evaluate the effectiveness of the proposed UnGSL strategy. 
Our experiments aim to answer the following research questions:
\begin{itemize}[leftmargin=1.0em]
    \item \textbf{RQ1:} Does the integration of UnGSL into existing GSL methods lead to performance improvements?
    \item \textbf{RQ2:} What is the impact of key configurations (\eg node-wise thresholds, asymmetric graphs, hyperparameters $\beta$ and $\tau$) on UnGSL performance? 
    \item  \textbf{RQ3:} Can UnGSL enhance the robustness of GNNs against structural noise, feature noise and lable noise? 
    \item \textbf{RQ4:} How well does UnGSL generalize across different GNN backbones? 
    \item \textbf{RQ5:} Does UnGSL introduce significant additional computational overhead? 
\end{itemize}
\begin{table*}[ht]
\renewcommand{\arraystretch}{1.20} 
\centering
\setlength{\abovecaptionskip}{0.cm}
\caption{Node classification accuracy±std comparsion(\%). 
Each experiment is repeated 10 times with different random seeds.
"OOM" denotes out of memory.
The top-performing results are marked in \textbf{bold}.}
\label{Tab:performance}
\setlength{\tabcolsep}{0.015\textwidth}{
{
\begin{tabular}{cccccccc}
\hline
        Model & Cora & Citeseer & Pubmed & BlogCatalog & Roman-empire & Flickr & Ogbn-arxiv \\ \hline
      
        GRCN & 84.70±0.31 & 72.49±0.77 & 78.94±0.16 & 76.17±0.23 & 44.29±0.28 & 59.55±0.31 & OOM  \\ 
        
        GRCN+UnGSL & \textbf{85.84±0.51} & \textbf{73.88±0.55} & \textbf{79.59±0.48} & 
        \textbf{76.78±0.11} & 
        \textbf{52.54±0.31} &\textbf{63.85±0.22} & OOM \\ 
        \hline

        PROGNN & 80.39±0.41 & 67.94±0.52 & OOM & 76.17±0.22 & OOM & 61.74±0.24 & OOM \\
        
        PROGNN+UnGSL & \textbf{81.86±0.55} & 
        \textbf{69.66±0.27} & 
        OOM & 
        \textbf{76.82±0.19} & 
        OOM &\textbf{61.92±0.43} & OOM \\ 
        \hline
        
        PROSE & {81.1±0.45} & {72.3±0.37} & 83.3±0.71 & {75.31±0.17} & 55.61±0.34  &
        59.77±1.13
        & 71.22±0.31\\ 
        
        PROSE+UnGSL & \textbf{81.90±0.31} & \textbf{73.10 ±0.32} & 
        \textbf{83.86±0.30} & \textbf{75.77±0.36} & 
        \textbf{56.17±0.31} 
        &
        \textbf{64.72±0.97}
        &
        \textbf{71.46±0.12}
        \\
        \hline
        IDGL & 84.50±0.5 & {72.49±0.67} & {82.83±0.33} & {89.66±0.28} & {46.67±0.56} & 85.77±0.16 & 70.45 ± 0.36 \\ 

        IDGL+UnGSL & \textbf{84.90±0.42} & \textbf{73.74±1.01} & \textbf{83.33±0.32} & \textbf{92.13±0.18} & \textbf{47.05±0.59} & \textbf{86.42±0.35} &\textbf{71.02±0.12}\\ 
        \hline
        
        SLAPS & 72.89±1.02 & {70.05±0.83} & {70.96±0.99} & {91.62±0.39} & {65.35±0.45} & 83.89±0.70 &  OOM \\

        SLAPS+UnGSL & \textbf{74.28±0.95} & \textbf{72.08±0.94} & \textbf{72.25±1.48} & \textbf{91.89±0.41} & \textbf{66.20±0.33} & \textbf{84.92±0.68} &OOM \\
        \hline

        SUBLIME & 83.30±1.04 & 72.36±0.68 & 80.41±0.69 & 95.20±0.21 & 63.48±0.53 & 88.68±0.23 & 71.37±0.13 \\
        
        SUBLIME+UnGSL & \textbf{84.24±0.91} & 
        \textbf{74.34±0.73} & 
        \textbf{80.84±0.92} & \textbf{96.18±0.38} & 
        \textbf{65.45±0.32} & \textbf{89.23±0.20} & \textbf{71.82±0.15}  \\ 
                \hline
\end{tabular}}}
\vspace{0.pt}
\end{table*}
\begin{table}[!t]
\renewcommand{\arraystretch}{1.2}
\vspace{0pt}
\centering
\setlength{\abovecaptionskip}{0.cm}
\caption{Performance comparison between the CUR decomposition and UnGSL modules.}
\label{Tab:CUR Vs UnGSL}
\setlength{\tabcolsep}{0.005\textwidth}{
{
\begin{tabular}{cccccc}
\hline
        Model & Cora & Citeseer & Roman-empire \\ \hline
      
        GRCN+CUR & 84.89±0.22 & 73.35±0.46  & 44.04±0.28 \\ 
        
        GRCN+UnGSL & \textbf{85.84±0.51} & \textbf{73.88±0.55}  & 
        \textbf{52.54±0.31}  \\ 
        \hline

        IDGL+CUR & 84.73±0.23 & 72.93±0.85  & OOM \\ 
        
        IDGL+UnGSL & \textbf{84.90±0.42} & \textbf{73.74±1.01} & 
        \textbf{47.05±0.59}  \\ 
        \hline
\vspace{-20pt}
\end{tabular}}}
\end{table}
\subsection{Experiment Settings}
\subsubsection{Datasets.}
To comprehensively evaluate UnGSL's performance on node classification, we follow previous works~\cite{zhou2023opengsl, yu2021graph} and select $7$ commonly used datasets, including four homophilous citation datasets~\cite{sen2008collective,OGB} (Cora, Citeseer, Pubmed and Ogbn-arxiv) and three heterophilous datasets (Blogcatalog~\cite{huang2017label} , Roman-Empire~\cite{platonov2023critical} and Flickr \cite{huang2017label}). 
The chosen datasets cover a wide range of homophily levels and graph sizes, allowing us to demonstrate UnGSL's effectiveness under various conditions. 
For a fair comparison, we strictly follow the data split settings used in the newly proposed benchmark for GSL~\cite{zhou2023opengsl}. 
Detailed statistics of these datasets are provided in Appendix~\ref{dataset}.

\subsubsection{Baselines.} To demonstrate UnGSL's generalizability across different GSL models, we select $6$ state-of-the-art GSL algorithms corresponding to the chosen datasets as baselines,
including supervised models (GRCN \cite{wang2021graph}, PROGNN \cite{jin2020graph}, IDGL \cite{chen2020iterative}, PROSE \cite{wang2023prose}, SLAPS \cite{fatemi2021slaps}), and a self-supervised model (SUBLIME \cite{liu2022towards}).
Detailed information of these baselines is provided in Appendix \ref{baselines}.
For all models, we report the average performance and standard deviations of 10 runs with different random seeds.

\subsubsection{Configuration.}
With regard to hyperparameter settings, only the hyperparameter $\beta$ and the learning rate of the threshold $\varepsilon$ require fine-tuning.
The threshold $\varepsilon$ is consistently initialized as a random number in the range [0, 1] across various datasets. 
The hyperparameter $\tau$ is simply fixed as a constant value of 2. 
The hyperparameter $\beta$ 
can be quickly tuned using parameter search tools like Bayesian optimization within the range of [0.001,1].
We optimized the UnGSL using Adam optimizer, with the learning rate selected from range [0.0001, 0.01].
For detailed hyperparameter settings please refer to Appendix~\ref{appx:hyper}.

{For all baselines, we strictly adhere to their original settings for hyperparameter tuning to ensure that they attain best performance. 
All GSL methods are evaluated based on the performance of GNNs on downstream tasks when using the learned structure. 
We also consider cross-architecture scenarios in Subsection~\ref{RQ4}, where GSL training and downstream tasks use different GNN architectures.}

\subsection{Main Results (RQ1)}

\label{RQ1}
\subsubsection{Comparison to Vanilla GSL Models.}
Table \ref{Tab:performance} presents the experimental results of the UnGSL module applied to various GSL models. 
We can observe that: 1) UnGSL significantly improves the node classification accuracy for all GSL models across all datasets with an average increase of 2.07$\%$, achieving new state-of-the-art performance in the GSL literature.
The improvement brought by UnGSL is particularly pronounced on GRCN \cite{yu2021graph}, achieving an average improvement of 5.12$\%$.  
These results empirically demonstrate UnGSL's effectiveness in further denoising the learned graph structure from the perspective of uncertainty, resulting in more  accurate predictions.
2) For the self-supervised GSL model SUBLIME \cite{liu2022towards}, UnGSL continues to achieve higher accuracy by utilizing contrastive loss in Equation (\ref{eq13}) as a proxy of uncertainty estimation to guide structure refinement, resulting in an average improvement of 1.40$\%$.
\begin{table}
\renewcommand{\arraystretch}{1.0}
\vspace{0pt}
\centering
\setlength{\abovecaptionskip}{0.cm}
\caption{Ablation study on the UnGSL when integrating with GRCN model \cite{yu2021graph}.}
\label{Tab:Ablation}
\setlength{\tabcolsep}{0.005\textwidth}{
{
\begin{tabular}{cccccc}
\hline
        Method & Cora & Citeseer 
        & Roman-empire \\ \hline
      
        Fixed $\boldsymbol{\varepsilon}$ & 85.23±0.15 & 73.2±0.96 
        & 44.72±0.14 \\ 
        $\text{Symmetrize}\, {\hat{\mathbf{S}}}$  & {85.03±0.40} & {73.74±0.49} &{52.28±0.025}   \\ 
        UnGSL & \textbf{85.84±0.51} & \textbf{73.88±0.55} & 
        \textbf{52.54±0.31}  \\ 
        \hline
\end{tabular}}}
\vspace{0.5pt}
\end{table}

\begin{figure}
    \vspace{-5pt}
    \centering    \includegraphics[width=0.720\linewidth, trim = 45 20 50 10]{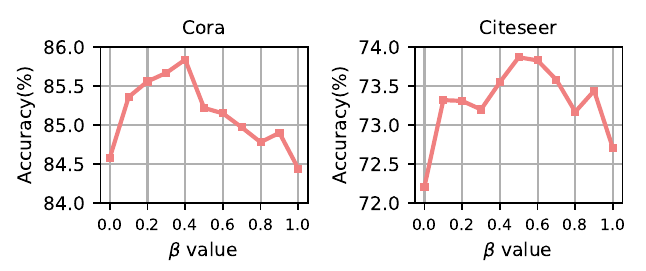}
    \caption{Comparsion of different $\beta$ on  Cora and Citeseer datasets.}
    \label{beta analysis}
    \vspace{-19pt}
    \Description{}
\end{figure}
\subsubsection{Comparison to the Label-oriented Directed GSL Module.}
\label{sec4.2.2}
Here we compare the performance of UnGSL with CUR decomposition \cite{lu2024latent}, another GSL module proposed recently to learn asymmetric graph structure by constructing directed edges from labeled nodes to unlabeled nodes.
Table \ref{Tab:CUR Vs UnGSL} shows the experiment results of UnGSL and CUR decomposition.
We can observe that: 1) UnGSL consistently outperforms CUR decomposition on supervised GSL methods.
2) UnGSL demonstrates better scalability with large-scale graphs compared to CUR decomposition. 
For example, IDGL+UnGSL is applicable to the Roman-Empire dataset and further enhances accuracy, while IDGL+CUR encounters an out-of-memory problem.

\subsection{Ablation Studies (RQ2)}

\label{RQ2}
\subsubsection{Effects of Adaptive Threshold $\boldsymbol{\varepsilon}$.}
To validate the effectiveness of the learnable threshold, we consider a variant where we fixed the $\boldsymbol{\varepsilon}$ in UnGSL during training phase.
Specifically, for each node, we select a fixed proportion of the most uncertain neighbors, as high-uncertainty neighbors. 
This proportion is consistent across all nodes. Next, we reweight edges by Equation (\ref{eq10}).
As shown in Table \ref{Tab:Ablation}, the learnable $\boldsymbol{\varepsilon}$ outperforms the fixed $\boldsymbol{\varepsilon}$ across 3 datasets.
The results indicate that the learnable threshold effectively differentiates high-uncertainty neighbors from low-uncertainty neighbors. 

\subsubsection{Superiority of Asymmetric Graph.}
To demonstrate the superiority of the asymmetric edges constructed by UnGSL, we symmetrize the graph refined by UnGSL during training phase, generating an symmetric graph structure. 
As shown in Table \ref{Tab:Ablation}, symmetrizing graph $\hat{\mathbf{S}}$ of UnGSL degrades its performance on node classification.
The underlying reason is that the symmetrization operation may perturb the graph by weakening edges from low-uncertainty neighbors and strengthening edge weights from high-uncertainty neighbors, which violates the core mechanism of UnGSL.

\subsubsection{Analysis on Hyperparameter $\beta$.}
To explore the role of $\beta$ in UnGSL's activation function $\psi( \cdot )$, we assign different values to  $\beta$ from the interval $[0,1]$ and evaluate the corresponding performance on the GRCN model \cite{yu2021graph}.
As shown in Figure \ref{beta analysis}, we can observe that:
(1) UnGSL enhances the accuracy of the GSL model by reducing edge weights with low-uncertainty neighbors using an appropriate $\beta$.
(2) UnGSL with positive $\beta$ consistently outperform UnGSL with $\beta=0$.
The results indicate that high-uncertainty neighbors still contain valuable information that enhances the node's representation, and removing connections from high-uncertainty nodes blindly may lead to information loss.
Results for additional GSL models are provided in Appendix \ref{appx: hyperbeta}.
\subsubsection{Analysis on Hyperparameter $\tau$.}
We  conduct additional experiments to evaluate the model's performance with different values of $\tau$
on Cora and Citeseer datasets using the base GSL model GRCN \cite{yu2021graph}.
The results are presented in Table \ref{Tab:tau}.
As can be seen, as $\tau$ increases, the performance initially improves and then declines. 
Hyperparameter $\tau$ controls influence of high-confidence neighbors. 
While fine-tuning $\tau$ could potentially enhance model performance, we find that simply setting $\tau = 2$ is sufficient to achieve good results.

\begin{table}
\renewcommand{\arraystretch}{1.2}
\centering
\setlength{\abovecaptionskip}{0.cm}
\caption{Comparsion of different $\tau$ on Cora and Citeseer datasets.}
\label{Tab:tau}
\setlength{\tabcolsep}{0.005\textwidth}{
{
\begin{tabular}{cccc}
\hline
        Method & $\tau =1 $ & $\tau=2$ 
        & $\tau=3$ \\ \hline
      
        Cora & 85.12±0.47 & 85.84±0.51 
        & 84.87±0.55 \\ 
        Citeseer  & {72.77±0.90} & {73.88±0.55} &{73.09±0.42}   \\  
        \hline
\end{tabular}}}
\vspace{-5pt}
\end{table}

\begin{table}[!t]
	\centering
        \small
        \setlength{\abovecaptionskip}{0.cm}
	\caption{Robustness analysis with random structural noise injection on Cora dataset.
 }
\setlength{\tabcolsep}{0.0011\textwidth}
	\label{tab:robust}
	
{
\begin{tabular}[t!]{c|ccccc|ccccc}
        \toprule
        {} &  \multicolumn{5}{c|}{Edge Deletion Level} &  \multicolumn{5}{c}{Edge Addition Level}\\
        \midrule
        {Methods} & 0$\%$ & 
        20$\%$ & 40$\%$ & 60$\%$ & 80$\%$ & 
        0$\%$ & 
        20$\%$ & 40$\%$ & 60$\%$ & 80$\%$\\
        \midrule
GRCN    & 84.70 & 83.00 & {79.87} & {78.47} & {75.17} & {84.70} & {78.03} & {75.13} & 73.20 & 69.43 \\
GRCN+UnGSL & {85.84} & {84.80} &{81.83} & {80.20} &{76.32} & {85.84} & {79.17} & {76.27} & {74.42} & {70.60}\\
Improve(\%) & {1.34\%} & {2.16\%} & {2.45\%} & {2.21\%} & {1.53\%} & {1.34\%} & {1.46\%} & {1.65\%} & 1.66\% & 1.69\% \\ 
\midrule
IDGL & 84.50 & 81.60 &80.31 & {76.57} & {72.18} & {84.50} & {78.59} & {77.24} & 74.43 & 73.11  \\
IDGL+UnGSL & {84.90} & {82.88} & {81.41} & {77.83} & {73.24} & {84.90} & {80.02} & {78.77} & {76.25} & {75.31} \\
Improve(\%) & 0.47\% & 1.57\% & 1.38\% & 1.65\% & 1.47\% & 0.47\% & 1.82\% & 1.98\% & 2.45\% & 3.01\% \\
\bottomrule
\end{tabular}
}
\vspace{-10pt}
\end{table}

\subsection{Robustness Analysis (RQ3)}
\label{RQ3}

To evaluate the robustness of UnGSL under structural noise, we compare the accuracy of the base GSL model and GSL+UnGSL under different noise levels and calculate the relative improvement achieved by GSL+UnGSL.
Specifically, we randomly add or remove edges from the original dataset.
We conduct experiments on the Cora dataset, and the results are presented in Table \ref{tab:robust}.
We can observe that:
1)  UnGSL consistently enhances the performance of GSL models across different noise levels.
2) With increasing levels of noise, UnGSL achieves greater relative improvements. 
For example, in the random edge addition scenario, IDGL+UnGSL achieves a 3.01\% improvement under 80\% edge perturbation, which is significantly higher than the 0.47\% improvement observed at the 0\% noise level.
In summary, the results suggest that UnGSL enhances robustness against structural noise.
We also validate UnGSL's robustness against feature noise and label noise, and the results are presented in Appendix \ref{appx:robust}.

\begin{table}[!t]
	\centering
        \small
        \setlength{\abovecaptionskip}{0.cm}
	\caption{Generalizability of UnGSL with different backbones on Cora dataset. 
 The value in bold signifies the top-performing result.}
	\label{tab:general}
	\resizebox{\linewidth}{!}
{
\begin{tabular}[t!]{c|cccc}
        \toprule
        {Methods} & SGC & 
        APPNP & GAT & JKNet\\
        \midrule
GRCN    & 84.40±0.00 & 84.00±0.15 &{81.45±1.04} & {83.73±1.33}  \\
GRCN+UnGSL & \textbf{84.80±0.10} & \textbf{84.40±0.61} &\textbf{82.45±1.04} & \textbf{84.53±1.38}\\
\midrule
IDGL & 78.00±0.00 & 82.13±0.32 &{79.87±1.01} & {76.23±1.19}\\
IDGL+UnGSL & \textbf{78.20±0.12} & \textbf{82.50±0.61} & \textbf{80.23±0.49} & \textbf{77.10±0.50}\\
\bottomrule
\end{tabular}
}
\vspace{-5pt}
\end{table}

\begin{table}
\renewcommand{\arraystretch}{1.0}
\centering
\setlength{\abovecaptionskip}{0.cm}
\caption{ Convergence time and GPU memory consumption of different models on Ogbn-arxiv dataset.}
\label{Tab:efficiencyarxiv}
\setlength{\tabcolsep}{0.005\textwidth}{
{
\begin{tabular}{cccccc}
\hline
        Model & Time(s) & Memory(MB) \\ \hline
      
        IDGL & 105 & 14,566  \\ 
        
        IDGL+UnGSL & 113 & 15,338   \\ 
        \hline

        SUBLIME & 807 & 14,586  \\ 
        
        SUBLIME+UnGSL & 886 & 14,678  \\ 
        \hline
\vspace{-15pt}
\end{tabular}}}
\end{table}

\subsection{Generalizability on GNN Models (RQ4)}
\label{RQ4}
We further consider the scenario where GSL training and downstream tasks use different GNN backbones. 
We evaluate the generalizability of the learned structures generated by the GSL+UnGSL on several other GNN models, including SGC \citep{wu2019simplifying}, APPNP \citep{gasteiger2018combining}, GAT \citep{veličković2018graph}, and JKNet \citep{xu2018representation}. 
The results are presented in Table \ref{tab:general}.
We observe that the graphs produced by GSL+UnGSL improve the prediction of various GNN models compared to those generated by vanilla GSL. 
Overall, UnGSL demonstrates its generalizability in enhancing performance across different GNN architectures.
Results for additional datasets are provided in Appendix \ref{appx: general}.

\subsection{Efficiency Analysis (RQ5)}
\label{seceffiency}
We analyze the time and memory efficiency of UnGSL on the Ogbn-Arxiv dataset, a large-scale graph with 169,343 nodes and 1,157,799 edges.
To assess time efficiency, we evaluate the algorithms by measuring the time taken to converge, \ie to reach optimal performance on the validation set. As shown in Table \ref{Tab:efficiencyarxiv}, UnGSL slightly increases convergence time (8.71\% on average) and training memory (2.97$\%$ on average) compared to the base GSL models.
The results demonstrate UnGSL's efficiency when applied to GSL models. 
\section{Related Work}
\label{sec5}
\subsection{Graph Neural Networks}
Graph neural networks (GNNs) are powerful models for learning node representations from graph data.
Existing GNNs can be categorized as spectral GNNs and spatial GNNs.
Spectral GNNs leverage eigenvectors and eigenvalues within the graph Laplacian matrix to design graph signal filters in the spectral domain \cite{bruna2013spectral,defferrard2016convolutional,li2021dimensionwise,he2022convolutional}.
Spatial GNNs \cite{kipf2017semisupervised,gilmer2017neural,veličković2018graph,wu2019simplifying,ding2022sketch,wu2021disenkgat} simplified spectral graph filter \cite{defferrard2016convolutional} using first-order approximation, which aggregates features from neighboring nodes in the spatial graph to generate node embeddings.
However, existing GNNs assume that the input graph structure is sufficiently clean for learning, whereas real-world graphs are often noisy and incomplete, which limits GNN performance on downstream tasks \cite{geisler2021robustness,chen2023bias}.
In this paper, we propose uncertainty-aware graph structure learning, which effectively denoises the graph structure to alleviate the above limitation.


\subsection{Graph Structure Learning}
Graph Structure Learning (GSL) aims to learn an optimal graph that improves the accuracy and robustness of Graph Neural Networks (GNNs) on downstream tasks.
Early GSL methods \cite{franceschi2019learning,jin2020graph} directly treat the target adjacency matrix as learnable parameters, incurring substantial
computational overhead and optimization challenge.
Mainstream GSL methods \cite{chen2020iterative,yu2021graph,wang2023prose,in2024self} learn edge weights based on node-pair embedding similarities, employing various metrics such as cosine similarity \cite{wang2023prose,in2024self}, inner product \cite{yu2021graph} or neural networks \cite{liu2022compact}.
These methods aim to increase graph homophily and typically achieve state-of-the-art performance, as these metrics can capture nodes with similar semantics.
However, these embedding-based GSL methods suffer from two limitations:
First, they construct edges solely based on embedding similarities while neglecting node uncertainty, which may introduce inferior information that poison the target node's embedding. 
Although some works~\cite{liu2022compact, duan2024structural} incorporate predictive uncertainty in structure learning, they only use it for cross-view structure fusion without distinguishing nodes of varying uncertainty levels within a graph.
Second, embedding-based methods impose bidirectional edges between nodes, disregarding their unequal influence due to varying levels of uncertainty.
Recently, several methods propose constructing directed edges from labeled to unlabeled nodes to facilitate the propagation of supervision signals~\cite{song2022towards,song2024optimal,lu2024latent}.
However, these methods fail to learn reasonable
asymmetric connections between the vast majority of unlabeled nodes.
Different from the above works, we propose an uncertainty-aware structure learning (UnGSL) strategy that learns node-wise thresholds to differentiate low-uncertainty from high-uncertainty neighbors and adaptively refine the corresponding edges.

\subsection{Uncertainty in GNNs}
GNNs inevitably present uncertainty towards their predictions, leading to unstable and erroneous prediction results \cite{wang2024uncertainty}.
In recent years, uncertainty in GNNs has been widely researched to adapt various tasks, including out-of-distribution (OOD) detection \cite{zhao2020uncertainty,stadler2021graph,yu2023uncertainty}, trustworthy GNN learning \cite{wang2021confident,hsu2022makes}, and GNN modeling \cite{um2023confidencebased}.
However, these uncertainty-based GNNs assume that the input graph structure is clean and primarily focus on incorporating uncertainty into the model architecture.
In contrast, our proposed uncertainty-aware graph structure learning aims to refine the edge based on node uncertainty, assisting in improving graph quality. 
\section{Conclusion}
\label{sec6}
In this paper, we first conduct theoretical and empirical analyses to demonstrate the detrimental impact of neighbors with high uncertainty on GNN learning.
Building on this, we propose the UnGSL strategy, a lightweight plug-in module that integrates seamlessly with state-of-the-art GSL models and boosts performance with minimal extra computational overhead.
UnGSL learns node-wise thresholds to differentiate between low-uncertainty and high-uncertainty neighbors, and adaptively refines the graph based on each node's uncertainty level.
Experiments demonstrate that UnGSL consistently enhances the performance and robustness of  GSL models.
In the future, we plan to explore more effective uncertainty metrics to accurately identify uncertain nodes in graph structure learning.
\begin{acks}
This work is supported by the National Natural Science Foundation of China (62476244,62372399,62476245), Zhejiang Provincial Natural Science Foundation of China (Grant No: LTGG23F030005).
\end{acks}

\bibliographystyle{ACM-Reference-Format}
\bibliography{sample-base}


\begin{thebibliography}{51}


\ifx \showCODEN    \undefined \def \showCODEN     #1{\unskip}     \fi
\ifx \showDOI      \undefined \def \showDOI       #1{#1}\fi
\ifx \showISBNx    \undefined \def \showISBNx     #1{\unskip}     \fi
\ifx \showISBNxiii \undefined \def \showISBNxiii  #1{\unskip}     \fi
\ifx \showISSN     \undefined \def \showISSN      #1{\unskip}     \fi
\ifx \showLCCN     \undefined \def \showLCCN      #1{\unskip}     \fi
\ifx \shownote     \undefined \def \shownote      #1{#1}          \fi
\ifx \showarticletitle \undefined \def \showarticletitle #1{#1}   \fi
\ifx \showURL      \undefined \def \showURL       {\relax}        \fi
\providecommand\bibfield[2]{#2}
\providecommand\bibinfo[2]{#2}
\providecommand\natexlab[1]{#1}
\providecommand\showeprint[2][]{arXiv:#2}

\bibitem[Abdar et~al\mbox{.}(2021)]%
        {abdar2021review}
\bibfield{author}{\bibinfo{person}{Moloud Abdar}, \bibinfo{person}{Farhad Pourpanah}, \bibinfo{person}{Sadiq Hussain}, \bibinfo{person}{Dana Rezazadegan}, \bibinfo{person}{Li Liu}, \bibinfo{person}{Mohammad Ghavamzadeh}, \bibinfo{person}{Paul Fieguth}, \bibinfo{person}{Xiaochun Cao}, \bibinfo{person}{Abbas Khosravi}, \bibinfo{person}{U~Rajendra Acharya}, {et~al\mbox{.}}} \bibinfo{year}{2021}\natexlab{}.
\newblock \showarticletitle{A review of uncertainty quantification in deep learning: Techniques, applications and challenges}.
\newblock \bibinfo{journal}{\emph{Information Fusion}}  \bibinfo{volume}{76} (\bibinfo{year}{2021}), \bibinfo{pages}{243--297}.
\newblock


\bibitem[Bruna et~al\mbox{.}(2013)]%
        {bruna2013spectral}
\bibfield{author}{\bibinfo{person}{Joan Bruna}, \bibinfo{person}{Wojciech Zaremba}, \bibinfo{person}{Arthur Szlam}, {and} \bibinfo{person}{Yann LeCun}.} \bibinfo{year}{2013}\natexlab{}.
\newblock \showarticletitle{Spectral networks and locally connected networks on graphs}. In \bibinfo{booktitle}{\emph{2nd International Conference on Learning Representations, ICLR 2014}}. \bibinfo{publisher}{Curran Associates Inc.}, \bibinfo{address}{Red Hook; NY; United States}.
\newblock


\bibitem[Chen et~al\mbox{.}(2024a)]%
        {chen2024macro}
\bibfield{author}{\bibinfo{person}{Hao Chen}, \bibinfo{person}{Yuanchen Bei}, \bibinfo{person}{Qijie Shen}, \bibinfo{person}{Yue Xu}, \bibinfo{person}{Sheng Zhou}, \bibinfo{person}{Wenbing Huang}, \bibinfo{person}{Feiran Huang}, \bibinfo{person}{Senzhang Wang}, {and} \bibinfo{person}{Xiao Huang}.} \bibinfo{year}{2024}\natexlab{a}.
\newblock \showarticletitle{Macro graph neural networks for online billion-scale recommender systems}. In \bibinfo{booktitle}{\emph{Proceedings of the ACM on Web Conference 2024}}. \bibinfo{publisher}{Association for Computing Machinery}, \bibinfo{address}{New York, NY, USA}, \bibinfo{pages}{3598--3608}.
\newblock


\bibitem[Chen et~al\mbox{.}(2023)]%
        {chen2023bias}
\bibfield{author}{\bibinfo{person}{Jiawei Chen}, \bibinfo{person}{Hande Dong}, \bibinfo{person}{Xiang Wang}, \bibinfo{person}{Fuli Feng}, \bibinfo{person}{Meng Wang}, {and} \bibinfo{person}{Xiangnan He}.} \bibinfo{year}{2023}\natexlab{}.
\newblock \showarticletitle{Bias and debias in recommender system: A survey and future directions}.
\newblock \bibinfo{journal}{\emph{ACM Transactions on Information Systems}} \bibinfo{volume}{41}, \bibinfo{number}{3} (\bibinfo{year}{2023}), \bibinfo{pages}{1--39}.
\newblock


\bibitem[Chen et~al\mbox{.}(2024b)]%
        {chen2024sigformer}
\bibfield{author}{\bibinfo{person}{Sirui Chen}, \bibinfo{person}{Jiawei Chen}, \bibinfo{person}{Sheng Zhou}, \bibinfo{person}{Bohao Wang}, \bibinfo{person}{Shen Han}, \bibinfo{person}{Chanfei Su}, \bibinfo{person}{Yuqing Yuan}, {and} \bibinfo{person}{Can Wang}.} \bibinfo{year}{2024}\natexlab{b}.
\newblock \showarticletitle{SIGformer: Sign-aware Graph Transformer for Recommendation}. In \bibinfo{booktitle}{\emph{Proceedings of the 47th International ACM SIGIR Conference on Research and Development in Information Retrieval}}. \bibinfo{publisher}{Association for Computing Machinery}, \bibinfo{address}{New York, NY, USA}, \bibinfo{pages}{1274--1284}.
\newblock


\bibitem[Chen et~al\mbox{.}(2020)]%
        {chen2020iterative}
\bibfield{author}{\bibinfo{person}{Yu Chen}, \bibinfo{person}{Lingfei Wu}, {and} \bibinfo{person}{Mohammed Zaki}.} \bibinfo{year}{2020}\natexlab{}.
\newblock \showarticletitle{Iterative deep graph learning for graph neural networks: Better and robust node embeddings}.
\newblock \bibinfo{journal}{\emph{Advances in Neural Information Processing Systems}}  \bibinfo{volume}{33} (\bibinfo{year}{2020}), \bibinfo{pages}{19314--19326}.
\newblock


\bibitem[Csisz{\'a}r et~al\mbox{.}(2004)]%
        {csiszar2004information}
\bibfield{author}{\bibinfo{person}{Imre Csisz{\'a}r}, \bibinfo{person}{Paul~C Shields}, {et~al\mbox{.}}} \bibinfo{year}{2004}\natexlab{}.
\newblock \showarticletitle{Information theory and statistics: A tutorial}.
\newblock \bibinfo{journal}{\emph{Foundations and Trends{\textregistered} in Communications and Information Theory}} \bibinfo{volume}{1}, \bibinfo{number}{4} (\bibinfo{year}{2004}), \bibinfo{pages}{417--528}.
\newblock


\bibitem[Defferrard et~al\mbox{.}(2016)]%
        {defferrard2016convolutional}
\bibfield{author}{\bibinfo{person}{Micha{\"e}l Defferrard}, \bibinfo{person}{Xavier Bresson}, {and} \bibinfo{person}{Pierre Vandergheynst}.} \bibinfo{year}{2016}\natexlab{}.
\newblock \showarticletitle{Convolutional neural networks on graphs with fast localized spectral filtering}.
\newblock \bibinfo{journal}{\emph{Advances in Neural Information Processing Systems}}  \bibinfo{volume}{29} (\bibinfo{year}{2016}).
\newblock


\bibitem[Deng et~al\mbox{.}(2024)]%
        {deng2024polynormer}
\bibfield{author}{\bibinfo{person}{Chenhui Deng}, \bibinfo{person}{Zichao Yue}, {and} \bibinfo{person}{Zhiru Zhang}.} \bibinfo{year}{2024}\natexlab{}.
\newblock \showarticletitle{Polynormer: Polynomial-Expressive Graph Transformer in Linear Time}. In \bibinfo{booktitle}{\emph{The Twelfth International Conference on Learning Representations}}. \bibinfo{publisher}{Curran Associates Inc.}, \bibinfo{address}{Red Hook; NY; United States}.
\newblock
\urldef\tempurl%
\url{https://openreview.net/forum?id=hmv1LpNfXa}
\showURL{%
\tempurl}


\bibitem[Ding et~al\mbox{.}(2022)]%
        {ding2022sketch}
\bibfield{author}{\bibinfo{person}{Mucong Ding}, \bibinfo{person}{Tahseen Rabbani}, \bibinfo{person}{Bang An}, \bibinfo{person}{Evan Wang}, {and} \bibinfo{person}{Furong Huang}.} \bibinfo{year}{2022}\natexlab{}.
\newblock \showarticletitle{Sketch-GNN: Scalable graph neural networks with sublinear training complexity}.
\newblock \bibinfo{journal}{\emph{Advances in Neural Information Processing Systems}}  \bibinfo{volume}{35} (\bibinfo{year}{2022}), \bibinfo{pages}{2930--2943}.
\newblock


\bibitem[Dong et~al\mbox{.}(2021)]%
        {dong2021equivalence}
\bibfield{author}{\bibinfo{person}{Hande Dong}, \bibinfo{person}{Jiawei Chen}, \bibinfo{person}{Fuli Feng}, \bibinfo{person}{Xiangnan He}, \bibinfo{person}{Shuxian Bi}, \bibinfo{person}{Zhaolin Ding}, {and} \bibinfo{person}{Peng Cui}.} \bibinfo{year}{2021}\natexlab{}.
\newblock \showarticletitle{On the equivalence of decoupled graph convolution network and label propagation}. In \bibinfo{booktitle}{\emph{Proceedings of the Web Conference 2021}}. \bibinfo{publisher}{Association for Computing Machinery}, \bibinfo{address}{New York,NY,United States}, \bibinfo{pages}{3651--3662}.
\newblock


\bibitem[Duan et~al\mbox{.}(2024)]%
        {duan2024structural}
\bibfield{author}{\bibinfo{person}{Liang Duan}, \bibinfo{person}{Xiang Chen}, \bibinfo{person}{Wenjie Liu}, \bibinfo{person}{Daliang Liu}, \bibinfo{person}{Kun Yue}, {and} \bibinfo{person}{Angsheng Li}.} \bibinfo{year}{2024}\natexlab{}.
\newblock \showarticletitle{Structural Entropy Based Graph Structure Learning for Node Classification}. In \bibinfo{booktitle}{\emph{Proceedings of the AAAI Conference on Artificial Intelligence}}, Vol.~\bibinfo{volume}{38}. \bibinfo{publisher}{{AAAI} Press}, \bibinfo{address}{Washington DC,USA}, \bibinfo{pages}{8372--8379}.
\newblock


\bibitem[Fatemi et~al\mbox{.}(2021)]%
        {fatemi2021slaps}
\bibfield{author}{\bibinfo{person}{Bahare Fatemi}, \bibinfo{person}{Layla El~Asri}, {and} \bibinfo{person}{Seyed~Mehran Kazemi}.} \bibinfo{year}{2021}\natexlab{}.
\newblock \showarticletitle{Slaps: Self-supervision improves structure learning for graph neural networks}.
\newblock \bibinfo{journal}{\emph{Advances in Neural Information Processing Systems}}  \bibinfo{volume}{34} (\bibinfo{year}{2021}), \bibinfo{pages}{22667--22681}.
\newblock


\bibitem[Franceschi et~al\mbox{.}(2019)]%
        {franceschi2019learning}
\bibfield{author}{\bibinfo{person}{Luca Franceschi}, \bibinfo{person}{Mathias Niepert}, \bibinfo{person}{Massimiliano Pontil}, {and} \bibinfo{person}{Xiao He}.} \bibinfo{year}{2019}\natexlab{}.
\newblock \showarticletitle{Learning discrete structures for graph neural networks}. In \bibinfo{booktitle}{\emph{International Conference on Machine Learning}}. PMLR, \bibinfo{publisher}{Curran Associates Inc.}, \bibinfo{address}{Long Beach, California, USA}, \bibinfo{pages}{1972--1982}.
\newblock


\bibitem[Gasteiger et~al\mbox{.}(2019)]%
        {gasteiger2018combining}
\bibfield{author}{\bibinfo{person}{Johannes Gasteiger}, \bibinfo{person}{Aleksandar Bojchevski}, {and} \bibinfo{person}{Stephan Günnemann}.} \bibinfo{year}{2019}\natexlab{}.
\newblock \showarticletitle{Combining Neural Networks with Personalized PageRank for Classification on Graphs}. In \bibinfo{booktitle}{\emph{International Conference on Learning Representations}}. \bibinfo{publisher}{Curran Associates Inc.}, \bibinfo{address}{Red Hook; NY; United States}.
\newblock
\urldef\tempurl%
\url{https://openreview.net/forum?id=H1gL-2A9Ym}
\showURL{%
\tempurl}


\bibitem[Geisler et~al\mbox{.}(2021)]%
        {geisler2021robustness}
\bibfield{author}{\bibinfo{person}{Simon Geisler}, \bibinfo{person}{Tobias Schmidt}, \bibinfo{person}{Hakan {\c{S}}irin}, \bibinfo{person}{Daniel Z{\"u}gner}, \bibinfo{person}{Aleksandar Bojchevski}, {and} \bibinfo{person}{Stephan G{\"u}nnemann}.} \bibinfo{year}{2021}\natexlab{}.
\newblock \showarticletitle{Robustness of graph neural networks at scale}.
\newblock \bibinfo{journal}{\emph{Advances in Neural Information Processing Systems}}  \bibinfo{volume}{34} (\bibinfo{year}{2021}), \bibinfo{pages}{7637--7649}.
\newblock


\bibitem[Gilmer et~al\mbox{.}(2017)]%
        {gilmer2017neural}
\bibfield{author}{\bibinfo{person}{Justin Gilmer}, \bibinfo{person}{Samuel~S Schoenholz}, \bibinfo{person}{Patrick~F Riley}, \bibinfo{person}{Oriol Vinyals}, {and} \bibinfo{person}{George~E Dahl}.} \bibinfo{year}{2017}\natexlab{}.
\newblock \showarticletitle{Neural message passing for quantum chemistry}. In \bibinfo{booktitle}{\emph{International Conference on Machine Learning}}. PMLR, \bibinfo{publisher}{JMLR.org}, \bibinfo{address}{Sydney NSW Australia}, \bibinfo{pages}{1263--1272}.
\newblock


\bibitem[He et~al\mbox{.}(2022)]%
        {he2022convolutional}
\bibfield{author}{\bibinfo{person}{Mingguo He}, \bibinfo{person}{Zhewei Wei}, {and} \bibinfo{person}{Ji-Rong Wen}.} \bibinfo{year}{2022}\natexlab{}.
\newblock \showarticletitle{Convolutional neural networks on graphs with chebyshev approximation, revisited}.
\newblock \bibinfo{journal}{\emph{Advances in Neural Information Processing Systems}}  \bibinfo{volume}{35} (\bibinfo{year}{2022}), \bibinfo{pages}{7264--7276}.
\newblock


\bibitem[Hsu et~al\mbox{.}(2022)]%
        {hsu2022makes}
\bibfield{author}{\bibinfo{person}{Hans Hao-Hsun Hsu}, \bibinfo{person}{Yuesong Shen}, \bibinfo{person}{Christian Tomani}, {and} \bibinfo{person}{Daniel Cremers}.} \bibinfo{year}{2022}\natexlab{}.
\newblock \showarticletitle{What makes graph neural networks miscalibrated?}
\newblock \bibinfo{journal}{\emph{Advances in Neural Information Processing Systems}}  \bibinfo{volume}{35} (\bibinfo{year}{2022}), \bibinfo{pages}{13775--13786}.
\newblock


\bibitem[Hu et~al\mbox{.}(2020)]%
        {OGB}
\bibfield{author}{\bibinfo{person}{Weihua Hu}, \bibinfo{person}{Matthias Fey}, \bibinfo{person}{Marinka Zitnik}, \bibinfo{person}{Yuxiao Dong}, \bibinfo{person}{Hongyu Ren}, \bibinfo{person}{Bowen Liu}, \bibinfo{person}{Michele Catasta}, {and} \bibinfo{person}{Jure Leskovec}.} \bibinfo{year}{2020}\natexlab{}.
\newblock \showarticletitle{Open Graph Benchmark: Datasets for Machine Learning on Graphs}. In \bibinfo{booktitle}{\emph{Advances in Neural Information Processing Systems 33: Annual Conference on Neural Information Processing Systems 2020, NeurIPS 2020, December 6-12, 2020, virtual}}, \bibfield{editor}{\bibinfo{person}{Hugo Larochelle}, \bibinfo{person}{Marc'Aurelio Ranzato}, \bibinfo{person}{Raia Hadsell}, \bibinfo{person}{Maria{-}Florina Balcan}, {and} \bibinfo{person}{Hsuan{-}Tien Lin}} (Eds.). \bibinfo{publisher}{Curran Associates Inc.}, \bibinfo{address}{Red Hook; NY; United States}.
\newblock
\urldef\tempurl%
\url{https://proceedings.neurips.cc/paper/2020/hash/fb60d411a5c5b72b2e7d3527cfc84fd0-Abstract.html}
\showURL{%
\tempurl}


\bibitem[Huang et~al\mbox{.}(2017)]%
        {huang2017label}
\bibfield{author}{\bibinfo{person}{Xiao Huang}, \bibinfo{person}{Jundong Li}, {and} \bibinfo{person}{Xia Hu}.} \bibinfo{year}{2017}\natexlab{}.
\newblock \showarticletitle{Label informed attributed network embedding}. In \bibinfo{booktitle}{\emph{Proceedings of the tenth ACM International Conference on Web Search and Data Mining}}. \bibinfo{publisher}{Association for Computing Machinery}, \bibinfo{address}{New York, NY, USA}, \bibinfo{pages}{731--739}.
\newblock


\bibitem[In et~al\mbox{.}(2024)]%
        {in2024self}
\bibfield{author}{\bibinfo{person}{Yeonjun In}, \bibinfo{person}{Kanghoon Yoon}, \bibinfo{person}{Kibum Kim}, \bibinfo{person}{Kijung Shin}, {and} \bibinfo{person}{Chanyoung Park}.} \bibinfo{year}{2024}\natexlab{}.
\newblock \showarticletitle{Self-Guided Robust Graph Structure Refinement}. In \bibinfo{booktitle}{\emph{Proceedings of the ACM on Web Conference 2024}}. \bibinfo{publisher}{Association for Computing Machinery}, \bibinfo{address}{New York, NY, USA}, \bibinfo{pages}{697--708}.
\newblock


\bibitem[Jin et~al\mbox{.}(2020)]%
        {jin2020graph}
\bibfield{author}{\bibinfo{person}{Wei Jin}, \bibinfo{person}{Yao Ma}, \bibinfo{person}{Xiaorui Liu}, \bibinfo{person}{Xianfeng Tang}, \bibinfo{person}{Suhang Wang}, {and} \bibinfo{person}{Jiliang Tang}.} \bibinfo{year}{2020}\natexlab{}.
\newblock \showarticletitle{Graph structure learning for robust graph neural networks}. In \bibinfo{booktitle}{\emph{Proceedings of the 26th ACM SIGKDD International Conference on Knowledge Discovery \& Data mining}}. \bibinfo{publisher}{Association for Computing Machinery}, \bibinfo{address}{New York, NY, USA}, \bibinfo{pages}{66--74}.
\newblock


\bibitem[Kipf and Welling(2017)]%
        {kipf2017semisupervised}
\bibfield{author}{\bibinfo{person}{Thomas~N. Kipf} {and} \bibinfo{person}{Max Welling}.} \bibinfo{year}{2017}\natexlab{}.
\newblock \showarticletitle{Semi-Supervised Classification with Graph Convolutional Networks}. In \bibinfo{booktitle}{\emph{International Conference on Learning Representations}}. \bibinfo{publisher}{Curran Associates, Inc.}, \bibinfo{address}{Red Hook; NY; United States}.
\newblock
\urldef\tempurl%
\url{https://openreview.net/forum?id=SJU4ayYgl}
\showURL{%
\tempurl}


\bibitem[Li et~al\mbox{.}(2021)]%
        {li2021dimensionwise}
\bibfield{author}{\bibinfo{person}{Qimai Li}, \bibinfo{person}{Xiaotong Zhang}, \bibinfo{person}{Han Liu}, \bibinfo{person}{Quanyu Dai}, {and} \bibinfo{person}{Xiao-Ming Wu}.} \bibinfo{year}{2021}\natexlab{}.
\newblock \showarticletitle{Dimensionwise separable 2-D graph convolution for unsupervised and semi-supervised learning on graphs}. In \bibinfo{booktitle}{\emph{Proceedings of the 27th ACM SIGKDD Conference on Knowledge Discovery \& Data Mining}}. \bibinfo{publisher}{Association for Computing Machinery}, \bibinfo{address}{New York, NY, USA}, \bibinfo{pages}{953--963}.
\newblock


\bibitem[Li et~al\mbox{.}(2023)]%
        {li2023gslb}
\bibfield{author}{\bibinfo{person}{Zhixun Li}, \bibinfo{person}{Liang Wang}, \bibinfo{person}{Xin Sun}, \bibinfo{person}{Yifan Luo}, \bibinfo{person}{Yanqiao Zhu}, \bibinfo{person}{Dingshuo Chen}, \bibinfo{person}{Yingtao Luo}, \bibinfo{person}{Xiangxin Zhou}, \bibinfo{person}{Qiang Liu}, \bibinfo{person}{Shu Wu}, \bibinfo{person}{Liang Wang}, {and} \bibinfo{person}{Jeffrey~Xu Yu}.} \bibinfo{year}{2023}\natexlab{}.
\newblock \showarticletitle{{GSLB}: The Graph Structure Learning Benchmark}. In \bibinfo{booktitle}{\emph{Thirty-seventh Conference on Neural Information Processing Systems Datasets and Benchmarks Track}}. \bibinfo{publisher}{Curran Associates Inc.}, \bibinfo{address}{Red Hook; NY; United States}.
\newblock
\urldef\tempurl%
\url{https://openreview.net/forum?id=xT3i5GS3zU}
\showURL{%
\tempurl}


\bibitem[Liu et~al\mbox{.}(2024)]%
        {liu2024scalable}
\bibfield{author}{\bibinfo{person}{Juncheng Liu}, \bibinfo{person}{Bryan Hooi}, \bibinfo{person}{Kenji Kawaguchi}, \bibinfo{person}{Yiwei Wang}, \bibinfo{person}{Chaosheng Dong}, {and} \bibinfo{person}{Xiaokui Xiao}.} \bibinfo{year}{2024}\natexlab{}.
\newblock \showarticletitle{Scalable and Effective Implicit Graph Neural Networks on Large Graphs}. In \bibinfo{booktitle}{\emph{The Twelfth International Conference on Learning Representations}}. \bibinfo{publisher}{Curran Associates, Inc.}, \bibinfo{address}{Red Hook; NY; United States}.
\newblock
\urldef\tempurl%
\url{https://openreview.net/forum?id=QcMdPYBwTu}
\showURL{%
\tempurl}


\bibitem[Liu et~al\mbox{.}(2022a)]%
        {liu2022compact}
\bibfield{author}{\bibinfo{person}{Nian Liu}, \bibinfo{person}{Xiao Wang}, \bibinfo{person}{Lingfei Wu}, \bibinfo{person}{Yu Chen}, \bibinfo{person}{Xiaojie Guo}, {and} \bibinfo{person}{Chuan Shi}.} \bibinfo{year}{2022}\natexlab{a}.
\newblock \showarticletitle{Compact graph structure learning via mutual information compression}. In \bibinfo{booktitle}{\emph{Proceedings of the ACM Web Conference 2022}}. \bibinfo{publisher}{Association for Computing Machinery}, \bibinfo{address}{New York, NY, USA}, \bibinfo{pages}{1601--1610}.
\newblock


\bibitem[Liu et~al\mbox{.}(2022b)]%
        {liu2022towards}
\bibfield{author}{\bibinfo{person}{Yixin Liu}, \bibinfo{person}{Yu Zheng}, \bibinfo{person}{Daokun Zhang}, \bibinfo{person}{Hongxu Chen}, \bibinfo{person}{Hao Peng}, {and} \bibinfo{person}{Shirui Pan}.} \bibinfo{year}{2022}\natexlab{b}.
\newblock \showarticletitle{Towards unsupervised deep graph structure learning}. In \bibinfo{booktitle}{\emph{Proceedings of the ACM Web Conference 2022}}. \bibinfo{publisher}{Association for Computing Machinery}, \bibinfo{address}{New York, NY, USA}, \bibinfo{pages}{1392--1403}.
\newblock


\bibitem[Lu et~al\mbox{.}(2024)]%
        {lu2024latent}
\bibfield{author}{\bibinfo{person}{Jianglin Lu}, \bibinfo{person}{Yi Xu}, \bibinfo{person}{Huan Wang}, \bibinfo{person}{Yue Bai}, {and} \bibinfo{person}{Yun Fu}.} \bibinfo{year}{2024}\natexlab{}.
\newblock \showarticletitle{Latent graph inference with limited supervision}.
\newblock \bibinfo{journal}{\emph{Advances in Neural Information Processing Systems}}  \bibinfo{volume}{36} (\bibinfo{year}{2024}).
\newblock


\bibitem[Ma et~al\mbox{.}(2022)]%
        {ma2022is}
\bibfield{author}{\bibinfo{person}{Yao Ma}, \bibinfo{person}{Xiaorui Liu}, \bibinfo{person}{Neil Shah}, {and} \bibinfo{person}{Jiliang Tang}.} \bibinfo{year}{2022}\natexlab{}.
\newblock \showarticletitle{Is Homophily a Necessity for Graph Neural Networks?}. In \bibinfo{booktitle}{\emph{International Conference on Learning Representations}}. \bibinfo{publisher}{Curran Associates, Inc.}, \bibinfo{address}{Austria}.
\newblock
\urldef\tempurl%
\url{https://openreview.net/forum?id=ucASPPD9GKN}
\showURL{%
\tempurl}


\bibitem[Platonov et~al\mbox{.}(2023)]%
        {platonov2023critical}
\bibfield{author}{\bibinfo{person}{Oleg Platonov}, \bibinfo{person}{Denis Kuznedelev}, \bibinfo{person}{Michael Diskin}, \bibinfo{person}{Artem Babenko}, {and} \bibinfo{person}{Liudmila Prokhorenkova}.} \bibinfo{year}{2023}\natexlab{}.
\newblock \showarticletitle{A critical look at the evaluation of GNNs under heterophily: Are we really making progress?}. In \bibinfo{booktitle}{\emph{The Eleventh International Conference on Learning Representations}}. \bibinfo{publisher}{Curran Associates, Inc.}, \bibinfo{address}{Kigali, Rwanda}.
\newblock


\bibitem[Sen et~al\mbox{.}(2008)]%
        {sen2008collective}
\bibfield{author}{\bibinfo{person}{Prithviraj Sen}, \bibinfo{person}{Galileo Namata}, \bibinfo{person}{Mustafa Bilgic}, \bibinfo{person}{Lise Getoor}, \bibinfo{person}{Brian Galligher}, {and} \bibinfo{person}{Tina Eliassi-Rad}.} \bibinfo{year}{2008}\natexlab{}.
\newblock \showarticletitle{Collective classification in network data}.
\newblock \bibinfo{journal}{\emph{AI magazine}} \bibinfo{volume}{29}, \bibinfo{number}{3} (\bibinfo{year}{2008}), \bibinfo{pages}{93--93}.
\newblock


\bibitem[Song et~al\mbox{.}(2022)]%
        {song2022towards}
\bibfield{author}{\bibinfo{person}{Zixing Song}, \bibinfo{person}{Yifei Zhang}, {and} \bibinfo{person}{Irwin King}.} \bibinfo{year}{2022}\natexlab{}.
\newblock \showarticletitle{Towards an optimal asymmetric graph structure for robust semi-supervised node classification}. In \bibinfo{booktitle}{\emph{Proceedings of the 28th ACM SIGKDD Conference on Knowledge Discovery and Data Dining}}. \bibinfo{publisher}{Association for Computing Machinery}, \bibinfo{address}{New York, NY, USA}, \bibinfo{pages}{1656--1665}.
\newblock


\bibitem[Song et~al\mbox{.}(2024)]%
        {song2024optimal}
\bibfield{author}{\bibinfo{person}{Zixing Song}, \bibinfo{person}{Yifei Zhang}, {and} \bibinfo{person}{Irwin King}.} \bibinfo{year}{2024}\natexlab{}.
\newblock \showarticletitle{Optimal block-wise asymmetric graph construction for graph-based semi-supervised learning}.
\newblock \bibinfo{journal}{\emph{Advances in Neural Information Processing Systems}}  \bibinfo{volume}{36} (\bibinfo{year}{2024}).
\newblock


\bibitem[Stadler et~al\mbox{.}(2021)]%
        {stadler2021graph}
\bibfield{author}{\bibinfo{person}{Maximilian Stadler}, \bibinfo{person}{Bertrand Charpentier}, \bibinfo{person}{Simon Geisler}, \bibinfo{person}{Daniel Z{\"u}gner}, {and} \bibinfo{person}{Stephan G{\"u}nnemann}.} \bibinfo{year}{2021}\natexlab{}.
\newblock \showarticletitle{Graph posterior network: Bayesian predictive uncertainty for node classification}.
\newblock \bibinfo{journal}{\emph{Advances in Neural Information Processing Systems}}  \bibinfo{volume}{34} (\bibinfo{year}{2021}), \bibinfo{pages}{18033--18048}.
\newblock


\bibitem[Um et~al\mbox{.}(2023)]%
        {um2023confidencebased}
\bibfield{author}{\bibinfo{person}{Daeho Um}, \bibinfo{person}{Jiwoong Park}, \bibinfo{person}{Seulki Park}, {and} \bibinfo{person}{Jin young Choi}.} \bibinfo{year}{2023}\natexlab{}.
\newblock \showarticletitle{Confidence-Based Feature Imputation for Graphs with Partially Known Features}. In \bibinfo{booktitle}{\emph{The Eleventh International Conference on Learning Representations}}. \bibinfo{publisher}{Curran Associates, Inc.}, \bibinfo{address}{Red Hook; NY; United States}.
\newblock
\urldef\tempurl%
\url{https://openreview.net/forum?id=YPKBIILy-Kt}
\showURL{%
\tempurl}


\bibitem[Veličković et~al\mbox{.}(2018)]%
        {veličković2018graph}
\bibfield{author}{\bibinfo{person}{Petar Veličković}, \bibinfo{person}{Guillem Cucurull}, \bibinfo{person}{Arantxa Casanova}, \bibinfo{person}{Adriana Romero}, \bibinfo{person}{Pietro Liò}, {and} \bibinfo{person}{Yoshua Bengio}.} \bibinfo{year}{2018}\natexlab{}.
\newblock \showarticletitle{Graph Attention Networks}. In \bibinfo{booktitle}{\emph{International Conference on Learning Representations}}. \bibinfo{publisher}{Curran Associates, Inc.}, \bibinfo{address}{Red Hook; NY; United States}.
\newblock
\urldef\tempurl%
\url{https://openreview.net/forum?id=rJXMpikCZ}
\showURL{%
\tempurl}


\bibitem[Wang et~al\mbox{.}(2024a)]%
        {wang2024distributionally}
\bibfield{author}{\bibinfo{person}{Bohao Wang}, \bibinfo{person}{Jiawei Chen}, \bibinfo{person}{Changdong Li}, \bibinfo{person}{Sheng Zhou}, \bibinfo{person}{Qihao Shi}, \bibinfo{person}{Yang Gao}, \bibinfo{person}{Yan Feng}, \bibinfo{person}{Chun Chen}, {and} \bibinfo{person}{Can Wang}.} \bibinfo{year}{2024}\natexlab{a}.
\newblock \showarticletitle{Distributionally Robust Graph-based Recommendation System}. In \bibinfo{booktitle}{\emph{Proceedings of the ACM on Web Conference 2024}}. \bibinfo{publisher}{Association for Computing Machinery}, \bibinfo{address}{New York, NY, USA}, \bibinfo{pages}{3777--3788}.
\newblock


\bibitem[Wang et~al\mbox{.}(2024b)]%
        {wang2024uncertainty}
\bibfield{author}{\bibinfo{person}{Fangxin Wang}, \bibinfo{person}{Yuqing Liu}, \bibinfo{person}{Kay Liu}, \bibinfo{person}{Yibo Wang}, \bibinfo{person}{Sourav Medya}, {and} \bibinfo{person}{Philip~S. Yu}.} \bibinfo{year}{2024}\natexlab{b}.
\newblock \showarticletitle{Uncertainty in Graph Neural Networks: A Survey}.
\newblock \bibinfo{journal}{\emph{Transactions on Machine Learning Research}} (\bibinfo{year}{2024}).
\newblock
\showISSN{2835-8856}
\urldef\tempurl%
\url{https://openreview.net/forum?id=0e1Kn76HM1}
\showURL{%
\tempurl}


\bibitem[Wang et~al\mbox{.}(2023)]%
        {wang2023prose}
\bibfield{author}{\bibinfo{person}{Huizhao Wang}, \bibinfo{person}{Yao Fu}, \bibinfo{person}{Tao Yu}, \bibinfo{person}{Linghui Hu}, \bibinfo{person}{Weihao Jiang}, {and} \bibinfo{person}{Shiliang Pu}.} \bibinfo{year}{2023}\natexlab{}.
\newblock \showarticletitle{Prose: Graph structure learning via progressive strategy}. In \bibinfo{booktitle}{\emph{Proceedings of the 29th ACM SIGKDD Conference on Knowledge Discovery and Data Mining}}. \bibinfo{publisher}{Association for Computing Machinery}, \bibinfo{address}{New York, NY, USA}, \bibinfo{pages}{2337--2348}.
\newblock


\bibitem[Wang et~al\mbox{.}(2021b)]%
        {wang2021graph}
\bibfield{author}{\bibinfo{person}{Ruijia Wang}, \bibinfo{person}{Shuai Mou}, \bibinfo{person}{Xiao Wang}, \bibinfo{person}{Wanpeng Xiao}, \bibinfo{person}{Qi Ju}, \bibinfo{person}{Chuan Shi}, {and} \bibinfo{person}{Xing Xie}.} \bibinfo{year}{2021}\natexlab{b}.
\newblock \showarticletitle{Graph structure estimation neural networks}. In \bibinfo{booktitle}{\emph{Proceedings of the Web Conference 2021}}. \bibinfo{publisher}{Association for Computing Machinery}, \bibinfo{address}{New York, NY, USA}, \bibinfo{pages}{342--353}.
\newblock


\bibitem[Wang et~al\mbox{.}(2021a)]%
        {wang2021confident}
\bibfield{author}{\bibinfo{person}{Xiao Wang}, \bibinfo{person}{Hongrui Liu}, \bibinfo{person}{Chuan Shi}, {and} \bibinfo{person}{Cheng Yang}.} \bibinfo{year}{2021}\natexlab{a}.
\newblock \showarticletitle{Be confident! towards trustworthy graph neural networks via confidence calibration}.
\newblock \bibinfo{journal}{\emph{Advances in Neural Information Processing Systems}}  \bibinfo{volume}{34} (\bibinfo{year}{2021}), \bibinfo{pages}{23768--23779}.
\newblock


\bibitem[Wu et~al\mbox{.}(2019)]%
        {wu2019simplifying}
\bibfield{author}{\bibinfo{person}{Felix Wu}, \bibinfo{person}{Amauri Souza}, \bibinfo{person}{Tianyi Zhang}, \bibinfo{person}{Christopher Fifty}, \bibinfo{person}{Tao Yu}, {and} \bibinfo{person}{Kilian Weinberger}.} \bibinfo{year}{2019}\natexlab{}.
\newblock \showarticletitle{Simplifying graph convolutional networks}. In \bibinfo{booktitle}{\emph{International Conference on Machine Learning}}. PMLR, \bibinfo{publisher}{Curran Associates, Inc.}, \bibinfo{address}{Long Beach, California, USA}, \bibinfo{pages}{6861--6871}.
\newblock


\bibitem[Wu et~al\mbox{.}(2021)]%
        {wu2021disenkgat}
\bibfield{author}{\bibinfo{person}{Junkang Wu}, \bibinfo{person}{Wentao Shi}, \bibinfo{person}{Xuezhi Cao}, \bibinfo{person}{Jiawei Chen}, \bibinfo{person}{Wenqiang Lei}, \bibinfo{person}{Fuzheng Zhang}, \bibinfo{person}{Wei Wu}, {and} \bibinfo{person}{Xiangnan He}.} \bibinfo{year}{2021}\natexlab{}.
\newblock \showarticletitle{DisenKGAT: knowledge graph embedding with disentangled graph attention network}. In \bibinfo{booktitle}{\emph{Proceedings of the 30th ACM International Conference on Information \& Knowledge Management}}. \bibinfo{publisher}{Association for Computing Machinery}, \bibinfo{address}{New York,NY,United States}, \bibinfo{pages}{2140--2149}.
\newblock


\bibitem[Xu et~al\mbox{.}(2018)]%
        {xu2018representation}
\bibfield{author}{\bibinfo{person}{Keyulu Xu}, \bibinfo{person}{Chengtao Li}, \bibinfo{person}{Yonglong Tian}, \bibinfo{person}{Tomohiro Sonobe}, \bibinfo{person}{Ken-ichi Kawarabayashi}, {and} \bibinfo{person}{Stefanie Jegelka}.} \bibinfo{year}{2018}\natexlab{}.
\newblock \showarticletitle{Representation learning on graphs with jumping knowledge networks}. In \bibinfo{booktitle}{\emph{International Conference on Machine Learning}}. PMLR, \bibinfo{publisher}{Curran Associates, Inc.}, \bibinfo{address}{Stockholm, Sweden}, \bibinfo{pages}{5453--5462}.
\newblock


\bibitem[Yu et~al\mbox{.}(2021)]%
        {yu2021graph}
\bibfield{author}{\bibinfo{person}{Donghan Yu}, \bibinfo{person}{Ruohong Zhang}, \bibinfo{person}{Zhengbao Jiang}, \bibinfo{person}{Yuexin Wu}, {and} \bibinfo{person}{Yiming Yang}.} \bibinfo{year}{2021}\natexlab{}.
\newblock \showarticletitle{Graph-revised convolutional network}. In \bibinfo{booktitle}{\emph{Machine Learning and Knowledge Discovery in Databases: European Conference, ECML PKDD 2020, Ghent, Belgium, September 14--18, 2020, Proceedings, Part III}}. Springer, \bibinfo{publisher}{Springer-Verlag}, \bibinfo{address}{Berlin, Heidelberg}, \bibinfo{pages}{378--393}.
\newblock


\bibitem[Yu et~al\mbox{.}(2023)]%
        {yu2023uncertainty}
\bibfield{author}{\bibinfo{person}{Linlin Yu}, \bibinfo{person}{Yifei Lou}, {and} \bibinfo{person}{Feng Chen}.} \bibinfo{year}{2023}\natexlab{}.
\newblock \showarticletitle{Uncertainty-aware Graph-based Hyperspectral Image Classification}. In \bibinfo{booktitle}{\emph{The Twelfth International Conference on Learning Representations}}. \bibinfo{publisher}{OpenReview.net}, \bibinfo{address}{Vienna, Austria}.
\newblock


\bibitem[Zhang et~al\mbox{.}(2021)]%
        {zhang2021hierarchical}
\bibfield{author}{\bibinfo{person}{Zhen Zhang}, \bibinfo{person}{Jiajun Bu}, \bibinfo{person}{Martin Ester}, \bibinfo{person}{Jianfeng Zhang}, \bibinfo{person}{Zhao Li}, \bibinfo{person}{Chengwei Yao}, \bibinfo{person}{Huifen Dai}, \bibinfo{person}{Zhi Yu}, {and} \bibinfo{person}{Can Wang}.} \bibinfo{year}{2021}\natexlab{}.
\newblock \showarticletitle{Hierarchical multi-view graph pooling with structure learning}.
\newblock \bibinfo{journal}{\emph{IEEE Transactions on Knowledge and Data Engineering}} \bibinfo{volume}{35}, \bibinfo{number}{1} (\bibinfo{year}{2021}), \bibinfo{pages}{545--559}.
\newblock


\bibitem[Zhao et~al\mbox{.}(2020)]%
        {zhao2020uncertainty}
\bibfield{author}{\bibinfo{person}{Xujiang Zhao}, \bibinfo{person}{Feng Chen}, \bibinfo{person}{Shu Hu}, {and} \bibinfo{person}{Jin-Hee Cho}.} \bibinfo{year}{2020}\natexlab{}.
\newblock \showarticletitle{Uncertainty aware semi-supervised learning on graph data}.
\newblock \bibinfo{journal}{\emph{Advances in Neural Information Processing Systems}}  \bibinfo{volume}{33} (\bibinfo{year}{2020}), \bibinfo{pages}{12827--12836}.
\newblock


\bibitem[Zhou et~al\mbox{.}(2023)]%
        {zhou2023opengsl}
\bibfield{author}{\bibinfo{person}{Zhiyao Zhou}, \bibinfo{person}{Sheng Zhou}, \bibinfo{person}{Bochao Mao}, \bibinfo{person}{Xuanyi Zhou}, \bibinfo{person}{Jiawei Chen}, \bibinfo{person}{Qiaoyu Tan}, \bibinfo{person}{Daochen Zha}, \bibinfo{person}{Yan Feng}, \bibinfo{person}{Chun Chen}, {and} \bibinfo{person}{Can Wang}.} \bibinfo{year}{2023}\natexlab{}.
\newblock \showarticletitle{Open{GSL}: A Comprehensive Benchmark for Graph Structure Learning}. In \bibinfo{booktitle}{\emph{Thirty-seventh Conference on Neural Information Processing Systems Datasets and Benchmarks Track}}. \bibinfo{publisher}{Curran Associates Inc.}, \bibinfo{address}{Red Hook, NY, USA}.
\newblock
\urldef\tempurl%
\url{https://openreview.net/forum?id=yXLyhKvK4D}
\showURL{%
\tempurl}


\end{thebibliography}
\balance
\clearpage
\appendix
\section{Proof of Proposition \ref{proposition 1} }
\label{proof}
To prove Proposition \ref{proposition 1}, we first introduce the log-sum inequality \cite{csiszar2004information} below.

\begin{lemma}[Log-sum Inequality]
Let $a_1,...,a_n$ and $b_1,...,b_n$ be non-negative numbers. 
Denote the sum of all $a_i$s by a and the sum of all $b_i$s by b.
Then log-sum inequality states that
\begin{equation}
    \sum_{i=1}^{n}a_{i}\text{log}\frac{a_i}{b_i} \geq a\text{log}\frac{a}{b},
\end{equation}
\end{lemma}
\textit{Proof.}
$\forall v_i \in \mathcal{V}$, by define the  logit $o_i=(\sum_{v_j \in \mathcal{N}{(v_i)}}\hat{\mathbf{A}}_{ij}x_j)\mathbf{W}$ and logit $o_{i}^{\prime}=x_i\mathbf{W}$, where $\sum_{v_j \in \mathcal{N}{(v_i)}}\hat{\mathbf{A}}_{ij}=1$, we have:
\begin{equation}
    o_i = \sum_{v_j \in \mathcal{N}{(v_i)}}\hat{\mathbf{A}}_{ij}o_{j}^{\prime} ,
\end{equation}
then for predictive probability vector $p_i$ of $o_i$, $\forall v_j \in \mathcal{N}{(v_i)}$, let $p_j^{\prime}$ denote the classification probabilities of its initial features.
We have:
 \begin{align}
    p_i &= \frac{o_{i} +  \mathbf{1}_{K} }{\sum_{k=1}^{K}(o_{ik} + 1)}\nonumber\\
    &= \frac{\sum_{v_j \in \mathcal{N}{(v_i)}}\hat{\mathbf{A}}_{ij}(o_{j}^{\prime} +  \mathbf{1}_{K} )}
    {\sum_{k=1}^{K}(\sum_{v_j \in \mathcal{N}{(v_i)}}\hat{\mathbf{A}}_{ij}o_{jk}^{\prime} + 1)}\nonumber\\
    &=\frac{\sum_{v_j \in \mathcal{N}{(v_i)}}p_{j}^{\prime}
     (\hat{\mathbf{A}}_{ij}{\sum_{k=1}^{K}(o_{jk}^{\prime} + 1)}) }
    {\sum_{v_j \in \mathcal{N}{(v_i)}}\hat{\mathbf{A}}_{ij}\sum_{k=1}^{K}(o_{jk}^{\prime} + 1)},\nonumber\\
    &=\sum_{v_j \in \mathcal{N}{(v_i)}}{\eta_j}p_{j}^{\prime},
 \end{align}
where,$\sum_{v_j \in \mathcal{N}{(v_i)}}{\eta_j}=1$.
Now we focus the $l$-th element in the probability vector:
\begin{align}
    p_{il} & =  \sum_{v_j \in \mathcal{N}{(v_i)}}{\eta_j}p_{jl}^{\prime}\nonumber\\
    \Longrightarrow 
    p_{il}\text{log}p_{il}&=(\sum_{v_j \in \mathcal{N}{(v_i)}}{\eta_j}p_{jl}^{\prime})\text{log}(\sum_{v_j \in \mathcal{N}{(v_i)}}{\eta_j}p_{jl}^{\prime})\nonumber\\
    &=(\sum_{v_j \in \mathcal{N}{(v_i)}}{\eta_j}p_{jl}^{\prime})\text{log}(\frac{{\sum_{v_j \in \mathcal{N}{(v_i)}}\eta_j}p_{jl}^{\prime}}{\sum_{v_j \in \mathcal{N}{(v_i)}}{\eta_j}})\nonumber\\
    &\leq \sum_{v_j \in \mathcal{N}{(v_i)}}{\eta_j}p_{jl}^{\prime}\text{log}p_{jl}^{\prime}\nonumber\\
    \Longrightarrow 
    -\sum_{l=1}^{K}p_{il}\text{log}p_{il} &\geq 
    -\sum_{v_j \in \mathcal{N}{(v_i)}}{\eta_j}\sum_{l=1}^{K}p_{jl}^{\prime}\text{log}p_{jl}^{\prime}\nonumber\\
    &=\sum_{v_j \in \mathcal{N}{(v_i)}}{\eta_j} u_{j}^{\prime},
\end{align}
which completes the proof.

\section{Datasets}
\label{dataset}
Table \ref{Tab:dataset stastics} shows the statistics of 7 datasets.
\section{Baselines}
\label{baselines}
we introduce all GSL models used in the experiment in this section.
\begin{itemize}[leftmargin=1.0em]
    \item {GRCN} \cite{yu2021graph} uses a GCN to extract topological features and compute the similarity between nodes as the edge weights of the learned graph.
    \item {ProGNN} \cite{jin2020graph} treats the graph as a learnable adjacency matrix and optimizes the sparsity, low-rankness, and feature smoothness of the graph structure.
    \item {PROSE} \cite{wang2023prose} identifies influential nodes using PageRank scores and reconstructs the graph structure by connecting these influential nodes.
    \item {IDGL} \cite{chen2020iterative} iteratively learns the graph structure and node embeddings, and introduces a node-anchor message-passing paradigm to scale IDGL to large graphs.
    \item {SLAPS} \cite{fatemi2021slaps} proposes learning a denoising autoencoder to filter noisy edges in the graph structure.
    \item {CUR} \cite{lu2024latent} is a model-agnostic structure learning module, which proposes constructing unidirectional edges from unlabeled nodes to labeled nodes via CUR decomposition to facilitate the propagation of supervision signals to unlabeled nodes.
    \item {SUBLIME} \cite{liu2022towards} is an unsupervised GSL model that employs contrastive learning between the learned graph and an augmented graph to enhance the robustness of the graph structure.
\end{itemize}

\section{Additional Experimental Results}
\subsection{Visualization of Node Entropy and Average Neighbor Entropy }
\label{ADDC1}
\begin{figure}[ht]
    \subfigure[Citeseer dataset]{
       \centering
     \includegraphics[width=0.8\linewidth,trim = 35 20 -10 -5]{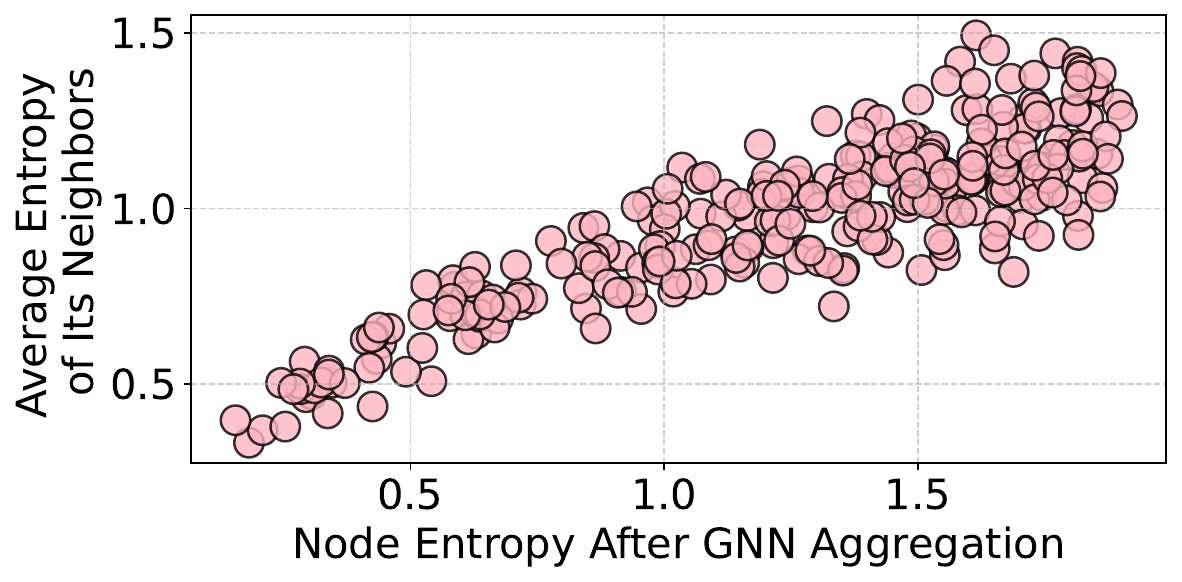}
     }
    \subfigure[Flickr dataset]{
        \centering
        \includegraphics[width=0.8\linewidth,trim = 35 20 -10 -5]{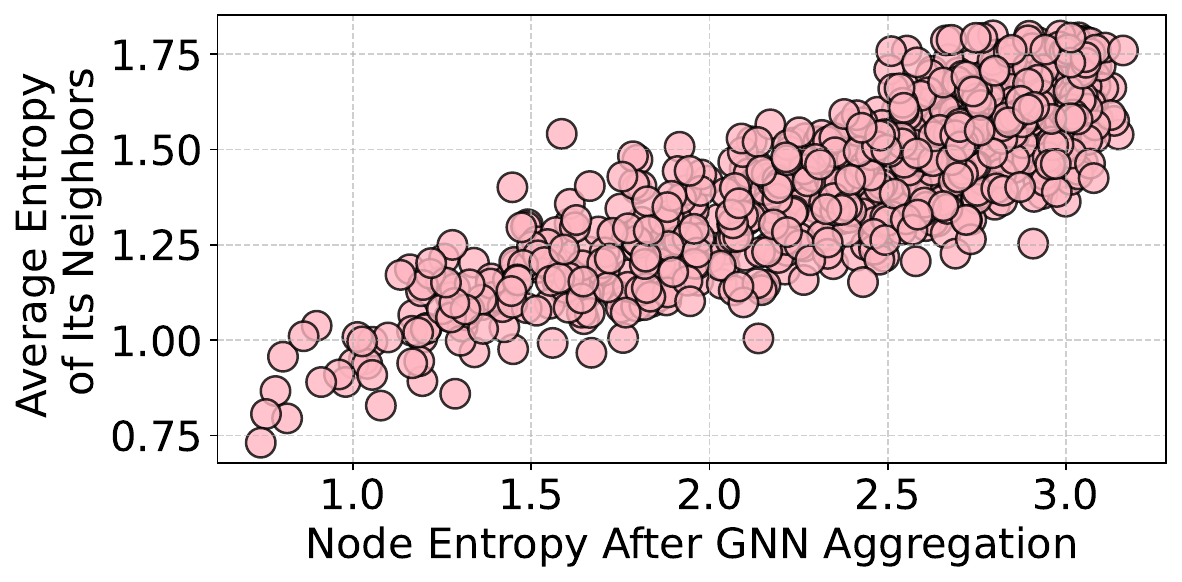}
            \Description{}
    }
    \caption{Visualization of node entropy after GNN aggregation  (i.e., $u_{i}$) alongside the average entropy of its neighbors  (i.e., $\sum_{v_j \in \mathcal{N}{(v_i)}}\mathbf{\hat{A}}_{ij}u_{j}^{\prime}$) on Citeseer and Flickr datasets.\label{Figuread2}}
        \Description{}
\end{figure}

We also conduct experiments on more datasets with different homophily characteristics.
Figure \ref{Figuread2} presents the node entropy after GNN aggregation, along with the average entropy of its neighbors, on the Citeseer and Flickr datasets.
We observe a strong positive correlation between the entropy of nodes and that of their neighbors across different datasets.


\subsection{Hyperparameter Settings}

\label{appx:hyper}
Table \ref{hyper} presents the hyperparameter settings of learning rate  and $\beta$ in UnGSL.

\subsection{Additional Results on the Hyperparameter \texorpdfstring{$\beta$}{}}
\label{appx: hyperbeta}
\begin{figure}[th]
    \centering    \includegraphics[width=0.70\linewidth, trim = 45 0 50 0]{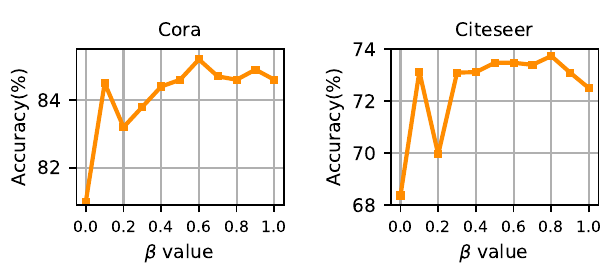}
    \caption{Comparsion of different $\beta$ on  Cora and Citeseer datasets.}
    \label{beta analysis IDGL}
    \vspace{-10pt}
        \Description{}
\end{figure}

We assign different values to $\beta$ from the interval [0,1] and evaluate the corresponding performance on the IDGL model \cite{chen2020iterative}. 
The results are presented in Figure \ref{beta analysis IDGL}.

\subsection{Additional Results for Robustness Analysis}
\label{appx:robust}
We evaluate the robustness of UnGSL against label noise and feature noise on the Cora dataset. The results are shown in Table \ref{tab:robust1}.
For feature noise, we randomly mask a proportion of node features by replacing their values with zeros. 
This approach allows us to investigate the performance of UnGSL when node features are subjected to varying degrees of damage.
For label noise, we randomly alter node labels.
The ratio of modifications  varies from 0 to 0.8 to simulate different levels of noise.
We can observe that:
1)  UnGSL consistently enhances the performance of GSL models across different noise levels.
2) With increasing levels of noise, UnGSL achieves greater relative improvements. 
For example, with 80\% wrong labels, GRCN+UnGSL achieves a 6.95\% improvement, which is significantly higher than the improvement under 0\% wrong labels scenario (1.34\%).
In summary, the results suggest that UnGSL enhances robustness to label noise and feature noise.

\subsection{Additional Results on Generalizability of UnGSL}
\label{appx: general}
Table \ref{tab:generalciteseer} illustrates the generalizability of UnGSL using various backbones on the Citeseer dataset.

\begin{table}[t]
\begin{minipage}[t]{\linewidth}
\renewcommand{\arraystretch}{1.1} 
\centering
\setlength{\abovecaptionskip}{-0.05cm}
\caption{Detailed statistics of node classification datasets.}
\label{Tab:dataset stastics}
\setlength{\tabcolsep}{0.008\textwidth}{
{
\begin{tabular}{lcccccc}
\hline
        Dataset & \#Nodes & \#Edges & \#Feat. & \#Avg.degree  & \#Homophily \\ \hline
      
        Cora & 2,708 & 5,278 & 1,433 & 3.9 &  0.81 \\ 
        
        Citeseer & {3,327} & {4,552} & {3,703} &  {2.7} &   0.74  \\ 

        Pubmed & 19,717 & 44,324 & 500 & 4.5 &  0.80  \\
        
        BlogCatalog & 5,196 & 
        171,743 & 
        8,189 & 66.1 & 
         0.40  \\ 
        Roman-Empire & 22,662 & 32,927 & 300 & 2.9 &  0.05  \\ 
        Flickr & 7,575 & 239,738 & 12,047 &  63.3 & 0.24
        \\
        Ogbn-arxiv & 169,343  & 1,157,799  &  767 & 13.67 & 0.65
        \\
        \hline
\end{tabular}}}
\end{minipage}%
\end{table}

\begin{table}[t]
\centering
\setlength{\abovecaptionskip}{-0.05cm}
\caption{Hyperparameter settings of learning rate, $\beta$ and initial value in UnGSL. }
\label{hyper}
{\small
\scalebox{0.73}{
\begin{tabular}{c|c|cccccc}
\hline
  & Setting & GRCN & PROGNN & PROSE & IDGL & SLAPS & SUBLIME\\
\midrule
\multirow{3}{*}{Cora} 
& lr         & 0.0005 & 0.0001 & 0.01   & 0.001   & 0.0001          & 0.001          \\
& $\beta$    & 0.40  &0.82 & 0.01   & 0.95 & 0.75          & 0.98          \\

\midrule
\multirow{3}{*}{Citeseer} 
&     lr     & 0.006 & 0.03 & 0.001 & 0.0005 & 0.03 & 0.03 \\
& $\beta$   & 0.56 & 0.57 & 0.75  & 0.78 &0.19          & 0.89          \\

\midrule
\multirow{3}{*}{Pubmed} 
& lr         & 0.09  & - & 0.01   & 0.001 &0.02         & 0.004          \\

& $\beta$   & 0.35  & - & 0.22 & 0.16 & 0.65 & 0.46          \\

\midrule
\multirow{2}{*}{BlogCatalog} 
& lr         & 0.0056  &0.0002 & 0.06  &0.01& 0.007          & 0.0001          \\

& $\beta$   & 0.65  & 0.87 & 0.36 & 0.90   & 0.18        & 0.47         \\

\midrule
\multirow{3}{*}{Roman-Empire} 
& lr         & 0.0008
 & - & 0.001 &0.002   & 0.02          & 0.007        \\

& $\beta$   & 0.001 & - & 0.001 & 0.82   & 0.83          & 0.54       \\

\midrule
\multirow{3}{*}{Flickr} 
& lr         &0.008
 & 0.0001 & 0.03 &0.002   & 0.036          & 0.0001        \\

& $\beta$   & 0.08 & 0.96 & 0.71 & 0.70   & 0.23          & 0.82       \\

\midrule
\multirow{3}{*}{Ogbn-arxiv} 
& lr         & -
 & - & 0.001 &0.003   & -          & 0.024        \\

& $\beta$   & - & - & 0.2 & 0.59  & -          & 0.48       \\

\bottomrule

\end{tabular}
}}
\label{tab:hype}
\vspace{-5pt}
\end{table}

\begin{table}[!t]
	\centering
        \small
        \setlength{\abovecaptionskip}{0.cm}
	\caption{Robustness analysis with random noise injection on Cora dataset.
    We evaluate the performance of UnGSL under feature noise and label noise.
 }
\setlength{\tabcolsep}{0.0011\textwidth}
	\label{tab:robust1}

{
\begin{tabular}[t!]{c|ccccc|ccccc}
        \toprule
        {} &  \multicolumn{5}{c|}{Feature Noise Level} & \multicolumn{5}{c}{Label Noise Level}\\
        \midrule
        {Methods} & 0$\%$ & 
        20$\%$ & 40$\%$ & 60$\%$ & 80$\%$ & 
        0$\%$ & 
        20$\%$ & 40$\%$ & 60$\%$ & 80$\%$
       \\
        \midrule
GRCN    &  84.70 & 82.47 &{81.60} & {79.40} & {72.40} & {84.70} & {81.33} & {73.32} & 65.21 & 58.13\\
GRCN+UnGSL & {85.84} & {83.70} & {82.80} & {80.62} & {73.70} & {85.84} & {83.43} & {74.52} & {67.94} & {62.17}\\
Improve(\%) & 1.34\% & 1.49\%  &{1.47\%} & {1.53\%} & {1.80\%} & {1.34\%} & {2.58\%} & {1.63\%} & 4.18\% & 6.95\%  \\ 
\midrule
IDGL & 84.50 & 82.10 &{79.25} & {75.23} & {65.16} & 84.50 &{77.60} & {70.80} & {54.81} & 43.65 \\
IDGL+UnGSL & {84.90} & {83.29} & {80.52} & {77.43} & {66.50} & {84.90} & {79.35} & {72.03} & {56.13} & {44.51} \\
Improve(\%) & 0.47\% & 1.45\% & 1.60\%  & 2.92\%  & 2.04\% & 0.47\%  & 2.26\% & {1.74\%} & {2.41\%} & {1.97\%} \\
\bottomrule
\end{tabular}
}
\end{table}

\begin{table}[!t]
	\centering
        \small
        \setlength{\abovecaptionskip}{-0.05cm}
	\caption{Generalizability of UnGSL with different backbones on Citeseer dataset. 
 The value in bold signifies the top-performing result.}
	\label{tab:generalciteseer}
	\resizebox{\linewidth}{!}
{
\begin{tabular}[ht]{c|cccc}
        \toprule
        {Methods} & SGC & 
        APPNP & GAT & JKNet \\
        \midrule
GRCN    & {72.00±0.00} & {72.73±0.96} & {70.77±1.06} & {71.83±0.72} \\
GRCN+UnGSL & \textbf{72.37±0.06} & \textbf{73.50±0.30} & \textbf{70.88±1.70} & \textbf{72.32±1.75}\\
\midrule
IDGL  & {71.90±0.00} & {72.20±0.75} & {63.83±0.97} & {71.00±1.15} \\
IDGL+UnGSL  & \textbf{73.40±0.00} & \textbf{73.37±0.61} & \textbf{67.00±0.69} & \textbf{72.33±0.9} \\
\bottomrule
\end{tabular}
}
\end{table}

\end{document}